\documentclass[a4paper]{article}
\usepackage[left=2.5cm,right=2.5cm,top=2cm,bottom=3cm]{geometry}
\usepackage[utf8]{inputenc}
\usepackage[numbers]{natbib}
\usepackage{soul}
\usepackage{color}
\usepackage{graphicx}
\usepackage{float}
\usepackage{subfigure}
\usepackage{url}
\usepackage{amsfonts}
\usepackage{booktabs}
\usepackage{siunitx}
\usepackage{authblk}
\usepackage{diagbox}
\usepackage{tabularx}

\providecommand{\keywords}[1]{
  \small	
  \textbf{\textit{Keywords}} #1
}

\begin{document}

\title{\hrule
\vspace{10pt}

    \Large{\textbf{Accelerated wind farm yaw and layout optimisation with multi-fidelity deep transfer learning wake models}}

\vspace{10pt}
\hrule}

\author[1,2]{\normalsize{Sokratis Anagnostopoulos\footnote{Corresponding Author, email: sokratis.anagnostopoulos19@imperial.ac.uk}}}
\author[1]{Jens Bauer}
\author[1,3]{Mariana  C. A. Clare}
\author[1]{Matthew D. Piggott}
\affil[1]{Department of Earth Science and Engineering, Imperial College London, UK}
\affil[2]{Mechanical Engineering, EPFL, Lausanne, Switzerland}
\affil[3]{ECMWF Bonn, Germany}

\maketitle

\begin{abstract}
Wind farm modelling has been an area of rapidly increasing interest with numerous analytical as well as computational-based approaches developed to extend the margins of wind farm efficiency and maximise power production. In this work, we present the novel ML framework wakeNet, which can reproduce generalised 2D turbine wake velocity fields at hub-height over a wide range of yaw angles, wind speeds and turbulence intensities (TIs), with a mean accuracy of 99.8\% compared to the solution calculated using the state-of-the-art wind farm modelling software FLORIS. As the generation of sufficient high-fidelity data for network training purposes can be cost-prohibitive, the utility of multi-fidelity transfer learning has also been investigated. Specifically, a network pre-trained on the low-fidelity Gaussian wake model is fine-tuned in order to obtain accurate wake results for the mid-fidelity Curl wake model. The robustness and overall performance of wakeNet on various wake steering control and layout optimisation scenarios has been validated through power-gain heatmaps, obtaining at least 90\% of the power gained through optimisation performed with FLORIS directly. We also demonstrate that when utilising the Curl model, wakeNet is able to provide similar power gains to FLORIS, two orders of magnitude faster (e.g. 10 minutes vs 36 hours per optimisation case). The wake evaluation time of wakeNet when trained on a high-fidelity CFD dataset is expected to be similar, thus further increasing computational time gains. These promising results show that generalised wake modelling with ML tools can be accurate enough to contribute towards active yaw and layout optimisation, while producing realistic optimised configurations at a fraction of the computational cost, hence making it feasible to perform real-time active yaw control as well as robust optimisation under uncertainty.
\end{abstract}


\section{Introduction}
    Renewable energy sources currently account for around 26\% of global energy production \cite{ren_report}. However, in order to reach Net Zero targets by 2050, the globally installed capacity for renewable energy generation must grow rapidly. In particular, the International Energy Agency (IEA) estimate that 390 GW of wind power must be added to the global energy grid every year until 2030, in order to reach Net Zero targets \cite{iea_report}. For context, in the record-breaking year of 2020, less than 100GW of wind power was added to the global grid. To meet these wind energy targets, we therefore not only need to build new wind farms, but also make existing farms more efficient. Wakes from upstream wind turbines can significantly decrease power production at downstream turbines due to reduced wind speeds and higher turbulent intensities. Over an entire wind farm, this effect can amount to a potential reduction of around 20\% in the annual energy production \cite{barthelmie2009modelling}, resulting in substantial economic losses \cite{lundquist2019costs}.
    
    In recent years, there have been extensive efforts to model turbine wakes in order to optimise wind farm configurations and improve the efficiency of existing wind farms through wake steering. Wake steering refers to the process of misaligning the upstream turbines to the direction of the wind (by changing the yaw angle) in order to steer the turbulent wake away from the downstream turbines. This has been shown to lead to overall wind farm power gains of up to 7\% \cite{dou2020optimization}. A variety of different analytical models exist for efficient wake modelling including the so-called Larsen \cite{larsen2009simple}, Jensen \cite{jensen1983note}, Curl \cite{Curl} and Gaussian \cite{Gaussian} models, which have been incorporated into tools such as the FLORIS (FLOw Redirection and Induction in Steady-state) software package \cite{floris}. FLORIS computes steady-state wakes in wind farms using a highly simplified physical description of the flow physics and is widely used in wake steering and wind farm optimisation studies \cite[e.g.][]{cioffi2020steady,gao2020comparisons,simley2021evaluation}. Wind power outputs from analytical models have been shown to compare relatively well with high fidelity numerical models such as Reynolds-averaged Navier-Stokes (RANS) \cite{martinez2021curled} and Large Eddy Simulations (LES) \cite{Gaussian} (minimum mean absolute error of 10--16\%), which are multiple orders of magnitude slower to execute. Although analytic models are not able to capture a detailed representation of the turbine velocity deficit, practitioners often compromise on accuracy to achieve more practical computational costs. This is particularly true in layout optimisation where the model needs to be executed a large number of times, and further still in the case of wake steering where the best results are achieved if this is performed in real time.
    
    During the last two decades, the exponential advancement of CPU, as well as GPU architectures, has opened up the possibility of using neural networks to predict wakes, instead of the more traditional analytical or numerical based models. This is part of a more significant trend in using machine learning techniques to optimise and control wind generation in wind farms \cite{wang2020review}. A suitable neural network trained on high fidelity data has the potential to provide reliable results within seconds, predicting flow properties of a wind farm that would otherwise require orders of magnitude higher computational times. In our work, we apply both fully connected neural networks (FCNNs) and convolutional neural networks (CNNs) to model turbine wakes. FCNNs have already been successfully used for wake representation by training on a large dataset of RANS simulation outputs to produce 3D wake profiles for wind turbines in a single row with varying wind speed and turbulence intensity (TI) \cite{ti2020wake}, by training on wakes from the analytical Jensen model to optimise wind farm layouts \cite{fischetti2019machine} and for wind farm optimisation based on generalised outputs with variable yaw, wind speed and turbulence \cite{anagnostopoulos2022offshore}. However, because wake representation can also be viewed as an image generation problem, in certain cases CNNs might be more appropriate due to their ability to accurately reproduce spatial features \cite{wang2020cnn}. CNNs have successfully been used to model flow fields around airfoils \cite{wu2020deep}, as well as to generate power response surface maps \cite{harrison2021machine}, where the CNN is trained on FLORIS outputs generated using the Gaussian analytical model. However, even though making predictions with neural networks is more efficient than traditional models, they still require large amounts of training data. For example, entire response surface maps have been generated in order to capture the power production for a given layout across all inflow conditions, requiring over 10 million training samples \cite{harrison2021machine}. Thus when generating appropriate training datasets, there is still a trade-off between cost and accuracy. More recent work has also investigated the use of multi-fidelity wake models to reduce the computational time of turbine wake evaluation \cite{LI2022116185,PAWAR2022867}.
    
    In this work, we seek to address the compromise between computational expense and high-fidelity accuracy by using a novel multi-fidelity approach with transfer learning (TL) where a new network is trained more efficiently using prior knowledge from a previous network \cite{tan2018survey}. This multi-fidelity transfer learning approach has been successfully used for solving PDEs and modelling flow past a wing \cite{chakraborty2021transfer,harada2022application}, but to the best of our knowledge this is the first time such an approach has been applied to the problem of wind energy optimisation. Specifically, we train our neural network first on wind field outputs generated with the simple Gaussian analytical model within FLORIS, and then use transfer learning with the more complex and more computationally expensive Curl model also implemented within FLORIS. In this way, we need to perform fewer simulations of the Curl model for the generation of training data and thus save computational time. In order to maximise the power generated by a wind farm, we must also estimate the power output from this wind field. A further novelty of this work is that instead of using an analytical or empirical approximation, we train a second neural network to predict the power and local TI for each turbine. Our efficient neural network framework can rapidly evaluate many different yaw angles and turbine location choices, and thus readily perform accurate optimisation in order to find the best possible power output.
    
    The aim of this work is thus to construct a neural network framework with multi-fidelity functionality for turbine wake modelling, capable of accurate active yaw control and layout optimisation for considerably lower computational costs than traditional models. The remainder of this work is structured as follows: initially we describe the wake models used for synthesising the training datasets and then we define the machine learning methods used in the wake modelling. We then present results which validate the adopted methodology and demonstrate promising computational gains which can contribute towards more efficient, generalised wake optimisation.
    
\section{Generating Training data}\label{sec:FLORIS}
In this work, we use the wind plant simulation and optimisation software FLORIS \cite{FLORIS_2021} to generate the wake fields used to train our machine learning framework. Within FLORIS, there are multiple models that can be used to simulate wakes and in this work we generate training data using both the Gaussian analytical model \cite{Gaussian} and the Curl model \cite{Curl}.

\subsection{Gaussian analytical model}
The Gaussian analytical model is based on the principle of mass and momentum conservation and assumes that the wake deficit is normally distributed \cite{Gaussian}. The normalised wake deficit is given by 
    \begin{equation}
        \frac{\Delta u(x, y, z)}{u_{0}} = \left(1 - \sqrt{1 - \frac{C_{t}}{8K}}\right)\exp\left(- \frac{1}{2K}\left(\left(\frac{z - z_{h}}{d_{0}}\right)^{2} + \left(\frac{y}{d_{0}}\right)^{2}\right)\right),
    \end{equation}
    where $\Delta u(x, y, z)$ is the wind speed wake deficit, $u_{0}$ the free-stream velocity, $C_{t}$ the thrust coefficient, $z_{h}$ the hub height, $d_{0}$ the rotor diameter and 
    \begin{equation}
        K = \left(k^{*} \frac{x}{d_{0}} + \epsilon\right)^{2},
    \end{equation}
    where $k^{*}$ defines the growth of the wake determined from experimental and LES data, and $\epsilon$ is the mass flow deficit rate at the rotor.

\subsection{Curl model}
   The Curl model is also included as part of FLORIS and is more complex and computationally expensive than the Gaussian analytical model. This model is derived by first considering the Reynolds-averaged Navier–Stokes (RANS) equation in the streamwise direction, 
    \begin{equation}
        u\frac{\partial u}{\partial x} + v\frac{\partial u}{\partial y} + w\frac{\partial u}{\partial z} = -\frac{1}{\rho}\frac{\partial p}{\partial x} + \nu\textsubscript{eff}\left(\frac{\partial^{2}u}{\partial x^{2}} + \frac{\partial^{2}u}{\partial y^{2}} + \frac{\partial^{2}u}{\partial z^{2}} \right),
    \end{equation}
    where all variables are time-averaged and $u$ is the streamwise velocity, $v$ the time-averaged spanwise velocity, $w$ the wall-normal velocity, $p$ the pressure, $\rho$ the fluid density and $\nu\textsubscript{eff}$ the effective viscosity. The velocities are then deconstructed into base components and perturbations, i.e. $u = U + u'$, where $(\cdot)'$ denotes the perturbation around the base solution component denoted by a capital letter. Linearising leads to 
        \begin{equation}
        U\frac{\partial u'}{\partial x} + V\frac{\partial (U + u')}{\partial y} + W\frac{\partial(U + u')}{\partial z} = -\frac{1}{\rho}\frac{\partial p}{\partial x} + \nu\textsubscript{eff}\left(\frac{\partial^{2}u'}{\partial x^{2}} + \frac{\partial^{2}(U + u')}{\partial y^{2}} + \frac{\partial^{2}(U + u')}{\partial z^{2}} \right),
    \end{equation}
    and making various simplifications including assuming the pressure gradient is zero (see \cite{Curl} for more details), we obtain the final simplified equation,
    \begin{equation}
        U\frac{\partial u'}{\partial x} + V\frac{\partial u'}{\partial y} + W \frac{\partial (U + u')}{\partial z} = \nu\textsubscript{eff}\left(\frac{\partial^{2}u'}{\partial x^{2}} + \frac{\partial^{2}u'}{\partial y^{2}} + \frac{\partial^{2}u'}{\partial z^{2}} \right).
    \end{equation}
    
    \noindent The latter is the equation solved by the Curl model to determine the wake deficit.
    
    \subsection{Dataset}
\begin{figure}[H]
        \centering
        \includegraphics[width=0.72\textwidth]{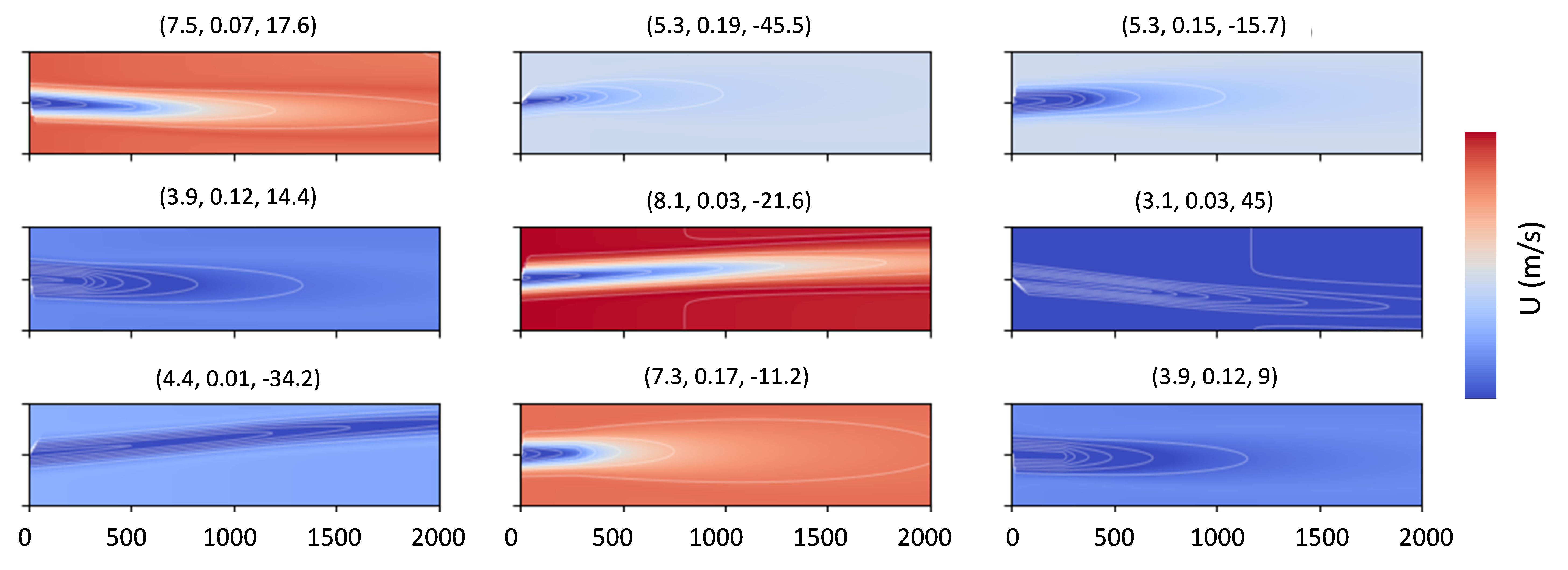}
        \caption{Indicative Gaussian wakes from the 2000 wake dataset with their corresponding input vector (wind speed, TI, yaw angle).}
        \label{fig:gaussian_wakes_dataset}
    \end{figure}

    The training process of the  core wake prediction Machine Learning module is based on a dataset of 2000 wakes generated using the Gaussian model and 2000 wakes generated using the Curl model. The range of initial wind speed, TI and yaw angle are [3, 15] m/s, [0.01, 0.2] and [-35, 35] degrees, respectively. The range of initial wind speeds is based on the power curve of the  NREL 5-MW wind turbine where the lower operational limit is 3 m/s and little further power gains from yaw optimisation can be made for wind speeds larger than 12 m/s \cite{KraghKnudHansen2015}. The TI limits are based on measured values \cite{chung_2002}. 
    Examples of the wakes generated using this method are shown in Figure \ref{fig:gaussian_wakes_dataset}. All 2000 wakes in the dataset were used for model training and 200 distinct wakes were used as a validation set. 
    
\section{Neural Network methods}
In this work, we introduce a novel machine learning framework called {\em WakeNet}, which expands on our previous model \cite{anagnostopoulos2022offshore} and aims at simulating multi-fidelity wake effects for yaw and layout optimisation. {\em WakeNet} is composed of three core neural network modules: a Deep Decoder Network (DDN) and a Convolutional Neural Network (CNN), which are trained to reproduce the downstream wake profile for a given turbine, and additional Fully Connected Neural Networks (FCCNs), which are used for the prediction of the local turbulence intensity (TI) and the generated power. The code is available at https://github.com/soanagno/wakenet.

\subsection{Wake profiles}\label{sec:wake_profiles}
\subsubsection{Deep Decoder Network (DDN)}
    
\begin{figure}[H]
        \includegraphics[scale=0.08]{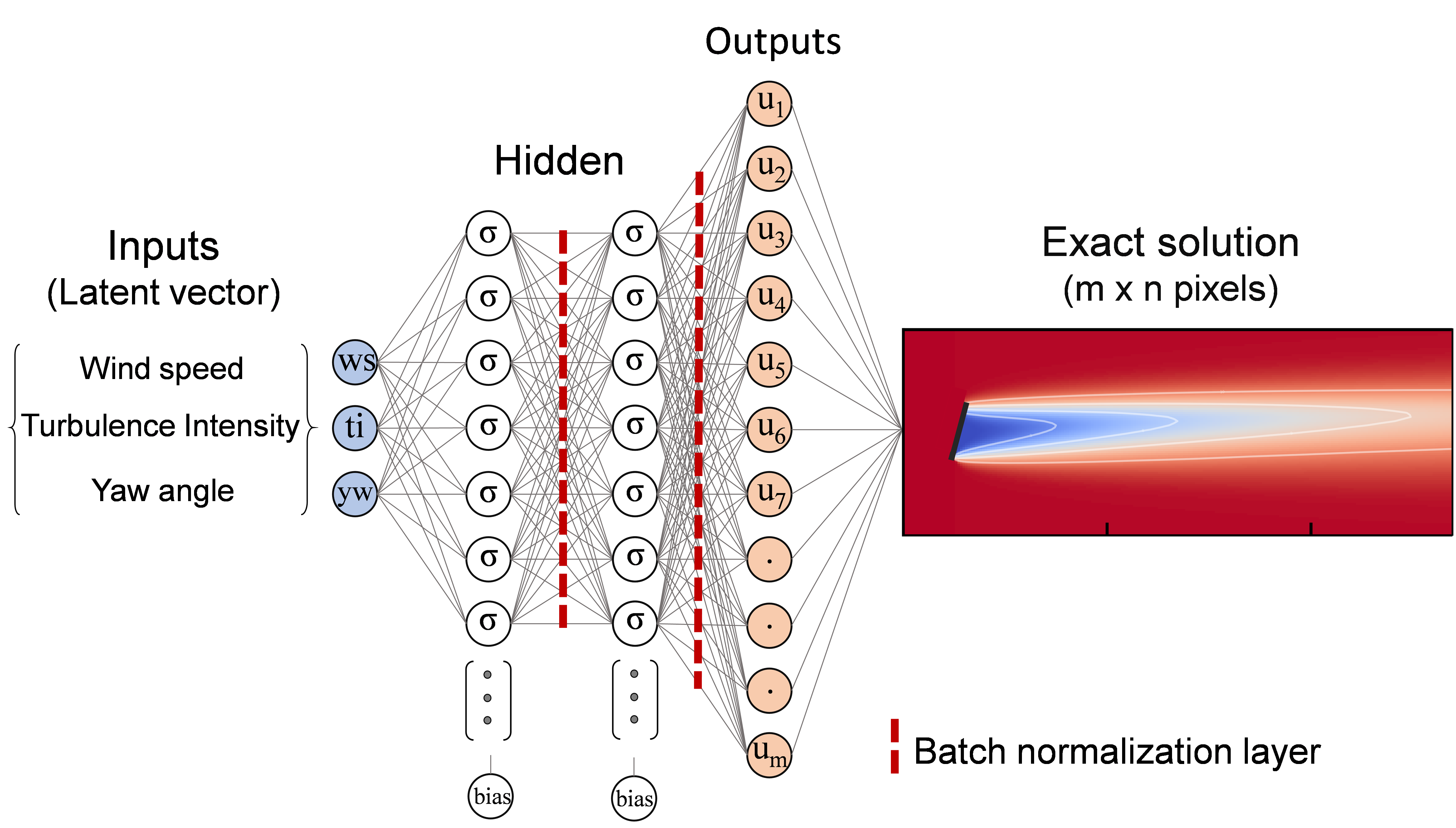}
        \centering
        \caption{Deep Decoder Network (DDN) architecture of the WakeNet module. The latent vector of wake parameters is decoded to the downstream velocity domain.}\label{fig:linear_arch}
    \end{figure}

    The first component in WakeNet is the fully-connected DDN. Its architecture is shown in Figure \ref{fig:linear_arch} and consists of at least two hidden layers of 200 neurons each, depending on the wake model, and outputs an $m$ by $n$ grid of pixels that represent the downstream velocity domain of the wake, while the input of the network is a vector of three wake parameters, namely the free-stream wind speed, the TI and the yaw angle of the turbine which have been normalised using mean/standard deviation normalisations. Additionally, a batch-normalisation layer has been applied after each hidden layer, because re-normalising the training batch in that manner significantly increases the performance of the trained network \cite{KingmaBa2014}. For the activation functions, we use $\tanh$ for the first two layers and a linear activation function for the output.
    
    The DDN is trained on a 2D horizontal slice from the 3D FLORIS wake outputs (the methods used to generate these outputs are described in Section \ref{sec:FLORIS}).  The advantage of this is that it makes it significantly faster to train the model. Moreover given that the training data comes from FLORIS which implicitly considers the effect of the sea/land surface on the atmospheric boundary layer, this means that, despite being a 2D slice, our output layer contains information about the velocity boundary layer of the site.
    
    We note further that the default resolution of FLORIS is a $200 \times 200$ grid, and thus we adopt this resolution for the output of the DDN. However, for higher resolution demands, the user can specify the desired resolution before training the model which will also produce an analytical wake dataset of that same resolution.
    
\subsubsection{Convolutional Neural Network (CNN)}

\begin{figure}[H]
        \includegraphics[scale=0.15]{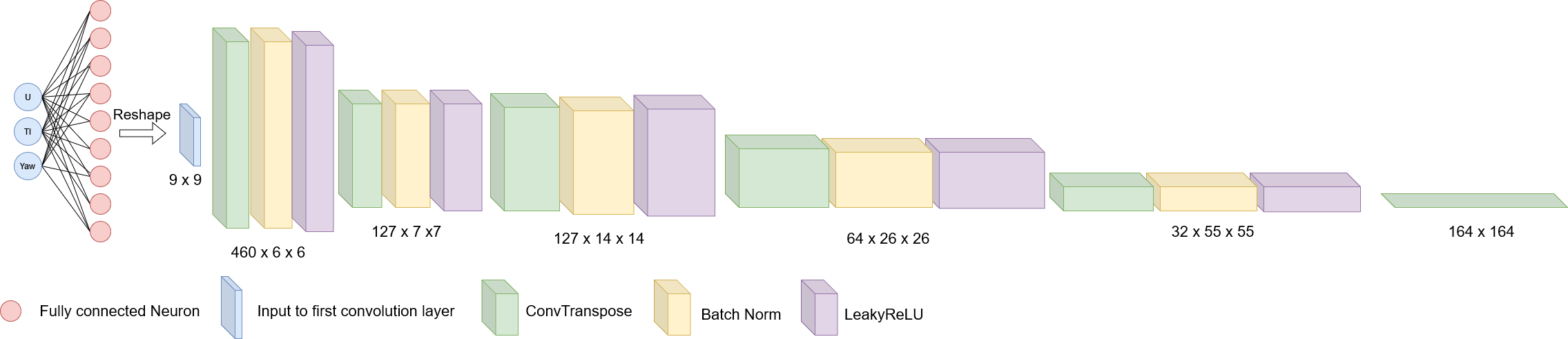}\label{fig:CNN}
        \centering
        \caption{Convolutional Neural Network (CNN) architecture.}
    \end{figure}

    In addition to the DDN, the WakeNet framework also incorporates a CNN architecture, which can also be deployed to generate the 2D wake velocity field. The CNN consists of a combination of fully connected and deconvolutional layers which reconstructs the flow field around a single wind turbine (Figure \ref{fig:CNN}). The first layer is a fully connected layer, for which the input is the wind speed, TI and yaw angle of the turbine at hub height. The output of this layer is then reshaped into a $3 \times 3$  array and passed to the deconvolutional layers. There are six deconvolutional layers, each consisting of ConvTranspose2d with leaky ReLU activation \cite{glaxsparse2011} and followed by batch normalisation. The output of the CNN model is a two-dimensional array with $200 \times 200$ pixels, which is a reconstruction of the flow field having the same dimensions as DDN. The number of pixels is set by the network architecture and training data, and thus can be easily changed. We note that the adopted CNN architecture was chosen using a hyperparameter search, where different layer configurations were tested.

\subsubsection{Wake superposition}

    The neural networks above provide a method to predict wakes. Most `analytical' wake models include an independent approach for the superposition of individual wakes in order to model an array of multiple wind turbines, as well as the interactions between them \cite{Gunn_2016, VogleWillden2020}. For this study, a superposition algorithm based on the sum of squares (SOS) model is deployed for the combination of multiple individual wakes produced. The superposition is modified with an approximate method of calculating a uniform velocity at the hub of each turbine. The final domain represents a collage of the individual wakes that comprise the examined wind farm. The SOS model has the following form: 

    \begin{equation}
    u_{i} =  \left(1 - \sqrt{\sum_{j=0}^{n} \left(1 - \frac{u_{i,j}}{u_{\text{hub},i}}\right)^2}\right) u_{\infty},
    \end{equation}

    \noindent where $u_{i}$ is the wind speed of the combined wakes at turbine $i$, $u_{i,j}$ is the wind speed at turbine $i$ which is influenced by the wake generated by the turbine $j$. $u_{\text{hub},i}$ is the hub wind speed at turbine $j$, $n$ is the number of wakes impacting the location and $u_{\infty}$ is the wind velocity at the wind park inlet.

\subsection{TI and Power prediction networks} \label{sec:Ti and Power networks}

\begin{figure}[H]
        \includegraphics[scale=0.07]{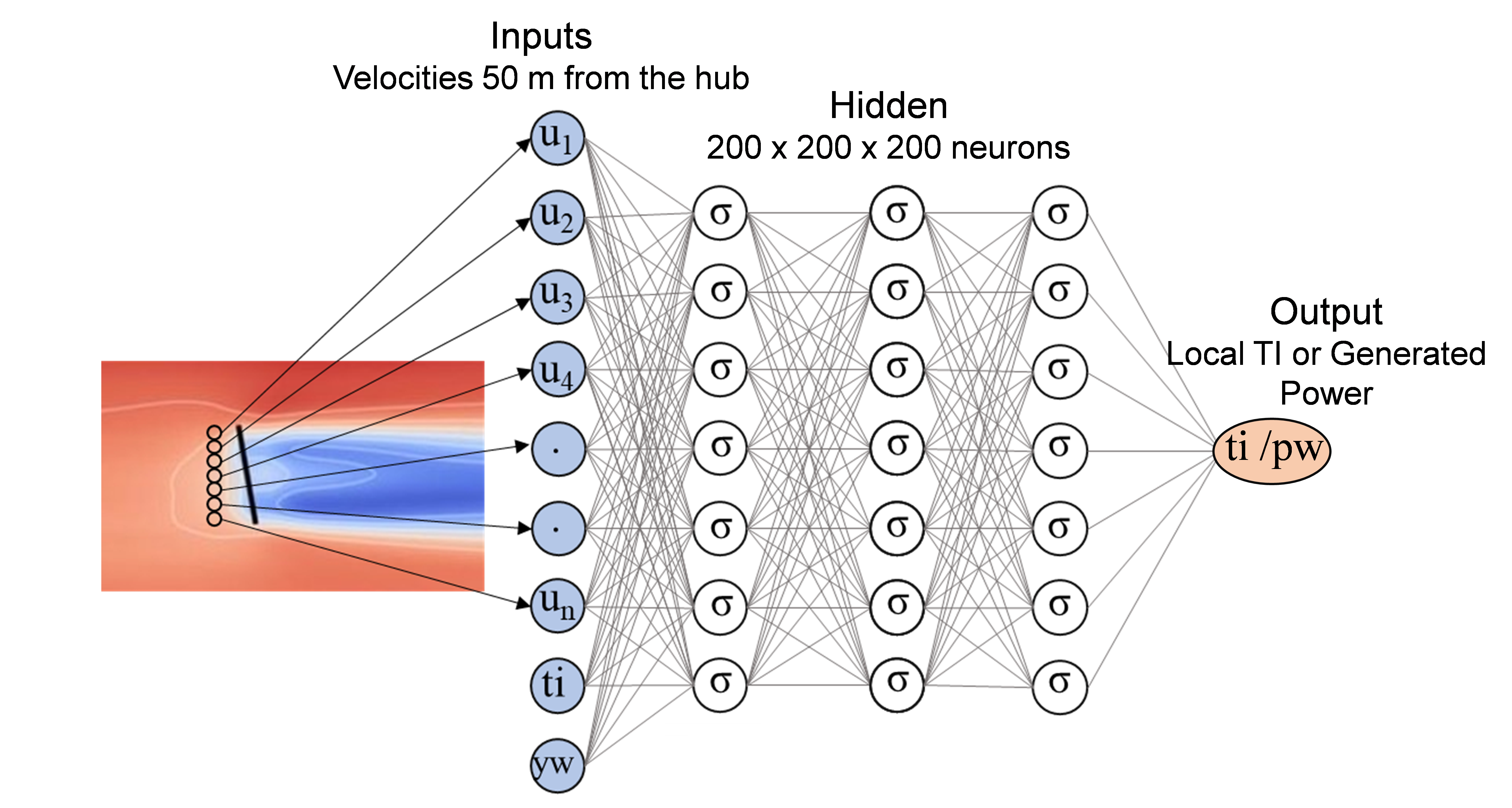}
        \centering
        \caption{TI and power predictor (FCNN) architecture.}\label{fig:FCNN}
    \end{figure}

    For the final component of the WakeNet framework, we predict the power and TI from the flow fields generated in Section \ref{sec:wake_profiles}. In the optimisation model, FLORIS, the power generated by a wind turbine is calculated from the three-dimensional flow field. However, the DDN and CNN outlined above generate a two-dimensional horizontal slice of the wind field, meaning that the power generated by the three-dimensional turbine needs to be predicted using two-dimensional data. To solve this problem, a FCNN is trained to predict the power output of a wind turbine from the wind speed along a horizontal line upstream the turbine (Figure \ref{fig:FCNN}). The TI of the flow field varies within the wind farm due to the turbulent wakes. Therefore, a second FCNN is used to predict the local TI at the turbines. As a result, through these FCNNs, it is possible to infer three-dimensional power and TI data from the two-dimensional flow field.
    The inputs to the TI and power predictor FCNNs are the wind speeds along a horizontal line that is 50 metres upstream the turbine, the inflow TI and turbine yaw angle. The wind speed line is parallel to the projected turbine rotor area and stretches along the whole diameter of the blades. The two FCNNs were trained on wind speeds produced by WakeNet and the corresponding FLORIS power output or local TI, respectively. The training data is generated by four example wind farms, consisting of six turbines. The power and local TI are calculated from the third-dimensional flow field using FLORIS while the extracted flow data is from a two-dimensional slice.
    
    \subsection{Training} \label{sec:training}
 For all neural networks used in the WakeNet framework, we take the standard approach of training by minimising the mean squared error (MSE) function using the Adam optimiser \cite{KingmaBa2014} to optimise this loss. The trainings are pefromed on an Nvidia GeForce RTX 3060 and require anywhere between 10 and 30 minutes, depending on the wake dataset size.
 
\begin{figure}[H]
        \centering
        \includegraphics[width=0.45\textwidth]{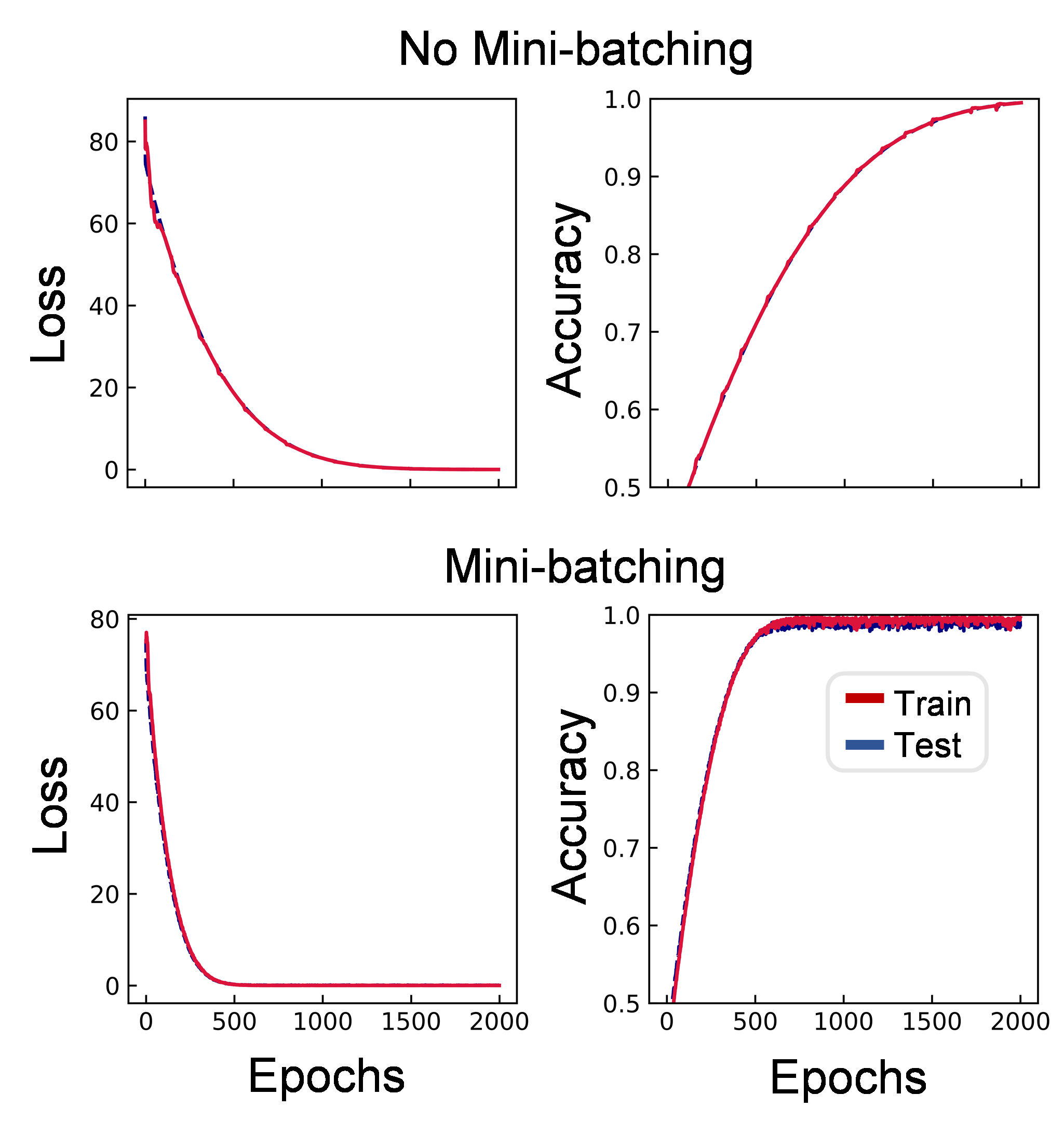}
        \caption{Training loss/accuracy curves with and without mini-batching.}
        \label{fig:4b}
    \end{figure}

 When training a neural network, there are a number of parameters that can be tuned to optimise training. We first consider the size of the mini-batch, which splits the training dataset into small batches used for each back-propagation step of the optimiser. Figure \ref{fig:4b} shows that a mini-batch size equal to 1/4 of the full dataset performs better than the full dataset, achieving higher training and validation accuracy within fewer epochs (2000 vs 500 epochs, respectively).
        
Another parameter that can be optimised is the learning rate of the optimiser. Smaller learning rates may require more epochs to train, but larger learning rates may result in the solution being `missed'. The optimal learning rate for both the DDN and CNN is 0.01. The optimal learning rate for the power predictor network is 0.003 for 150 epochs and for the TI predictor network is 0.0065 for 80 epochs. For both TI and power prediction networks, a learning rate scheduler was used to reduce the learning rate by a factor of 0.8, if the loss on the validation dataset did not decrease after 15 steps. This helped to further reduce the error on validation and test sets. 

\subsection{Multi-fidelity transfer learning}

    The ultimate goal of this work is to create a framework that can be deployed on any wind farm site to produce rapid and accurate optimised predictions for the yaw setting of the turbines. The higher the fidelity of the wake model used in the final training, the more accurate the prediction. However, producing a large wake dataset (at least 2000 wakes) of high-fidelity CFD results for any turbine type is not a viable option, as that would require immense computational resources. In this work we demonstrate the capabilities of multi-fidelity transfer learning between the Curl model (playing the role of our higher fidelity model) and the Gaussian wake model (a computationally cheaper model).

\begin{figure}[H]
        \centering
        \includegraphics[width=0.59\textwidth]{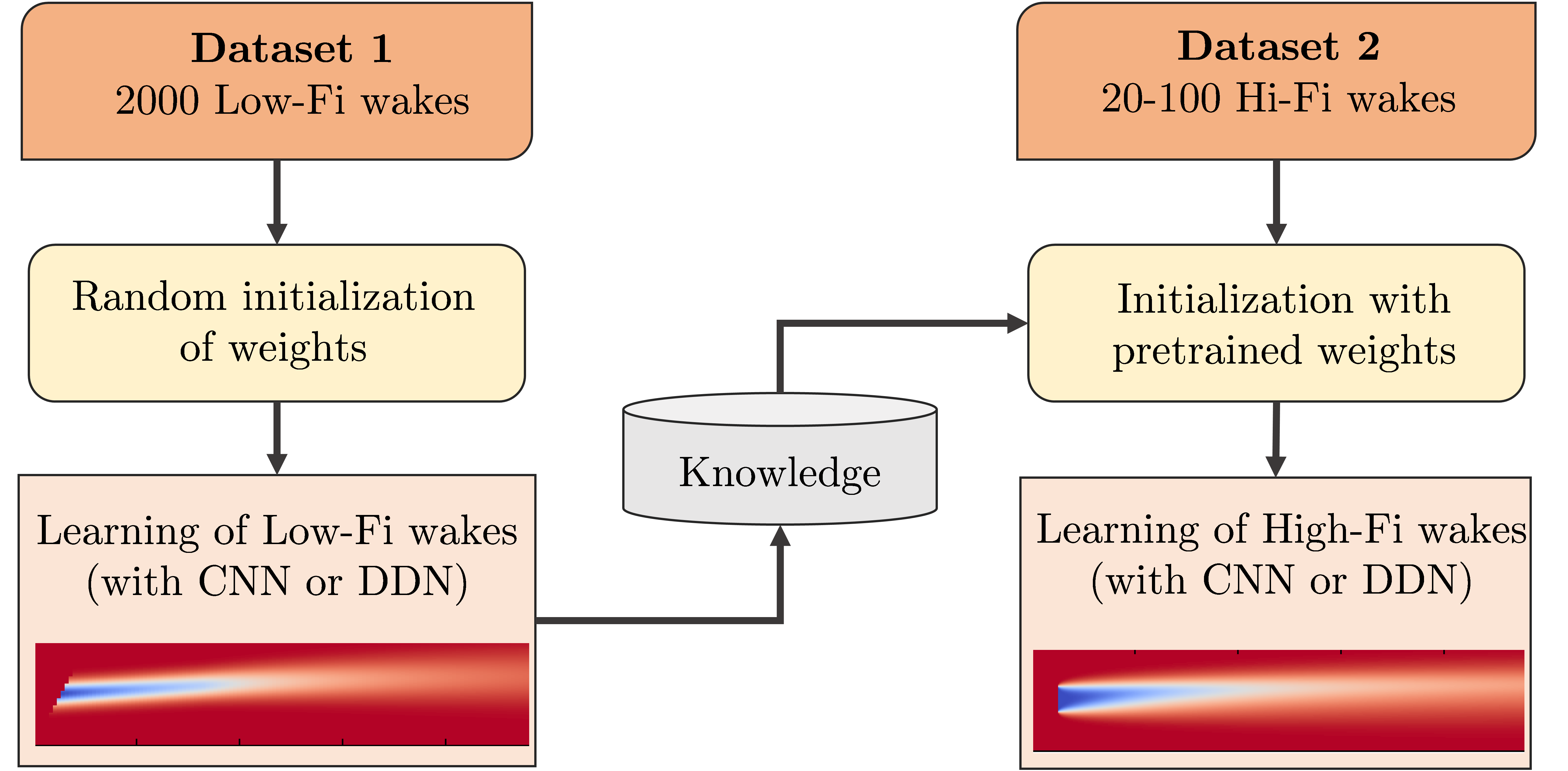}
        \caption{Multi-fidelity transfer learning process diagram.}
        \label{fig:Multi-fidelity diagram}
    \end{figure}

    Transfer learning models focus on leveraging the pre-gained knowledge obtained during training on a problem and applying it to a similar problem. One of the applications of this technique is to fine-tune a pre-trained neural network in order to produce accurate results when the dataset is limited. The main focus of this work is to transfer the knowledge from a low-fidelity computationally cheap wake model, such as Gaussian, to the higher-fidelity Curl model, as shown in Figure \ref{fig:Multi-fidelity diagram}. To perform the transfer learning, WakeNet is first trained on a dataset of 2000 wakes produced using the Gaussian model. The first two layers of the DNN are then \textquoteleft frozen\textquoteright, meaning their weights are unchanged by the back-propagation algorithm during training, whilst the last layer is trained based on the new information contained in the Curl dataset.
    
\section{Results}

    To demonstrate the capabilities of WakeNet, several indicative cases are examined both for single and multiple wakes. The metric of mean absolute error (\%) between FLORIS and each neural network submodule is used to assess the accuracy in evaluating the wake profile.
    
\subsection{Wake field comparison}
\begin{figure}[H]
        \centering
        \includegraphics[width=0.80\textwidth]{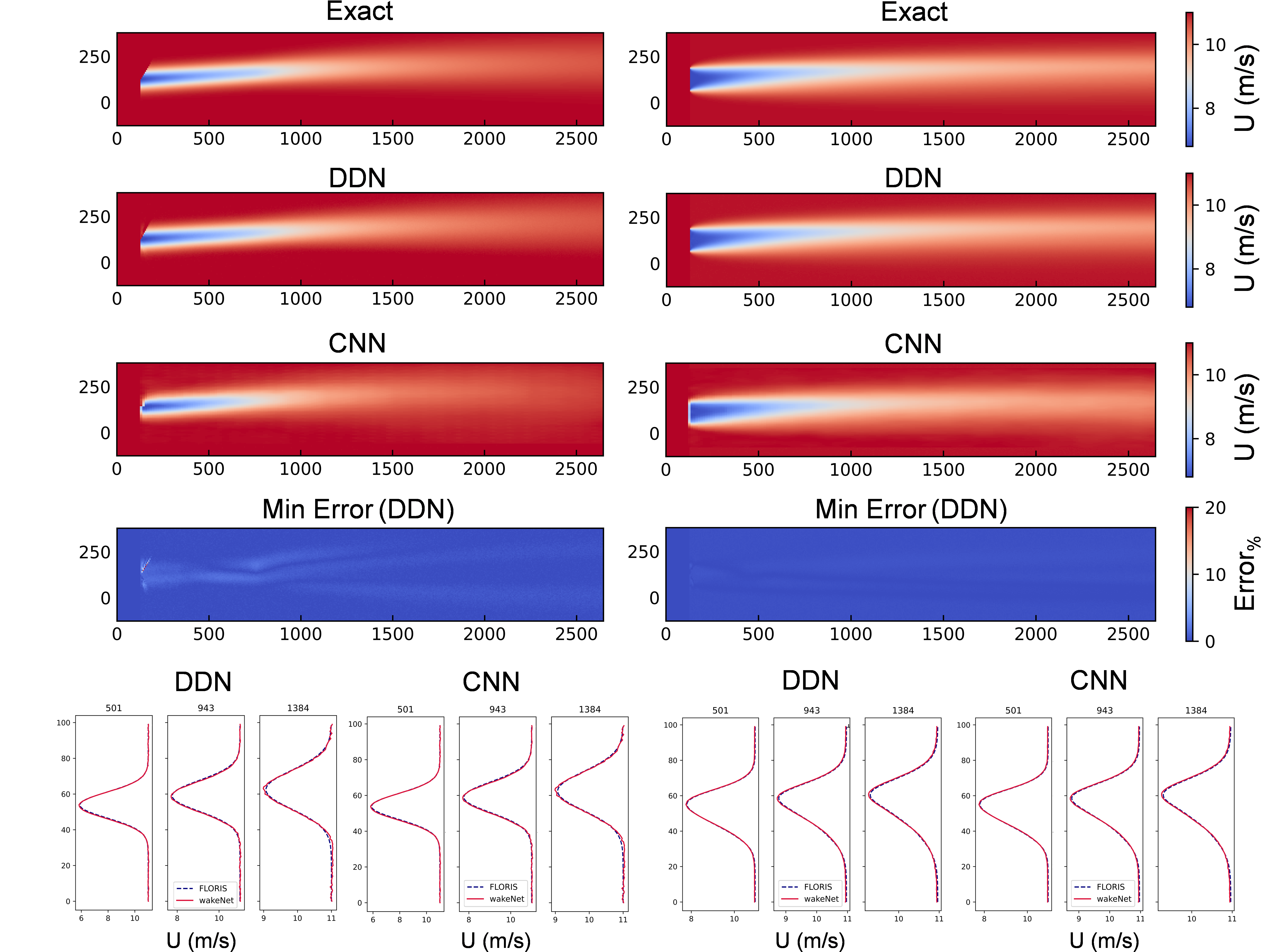}
        \caption{Example single wake comparison for Gaussian (left) and Curl (right) models. Exact refers to the results obtained using FLORIS.}
        \label{fig:Single_wake_results}
    \end{figure}

\begin{figure}[H]
        \centering
        \includegraphics[width=0.80\textwidth]{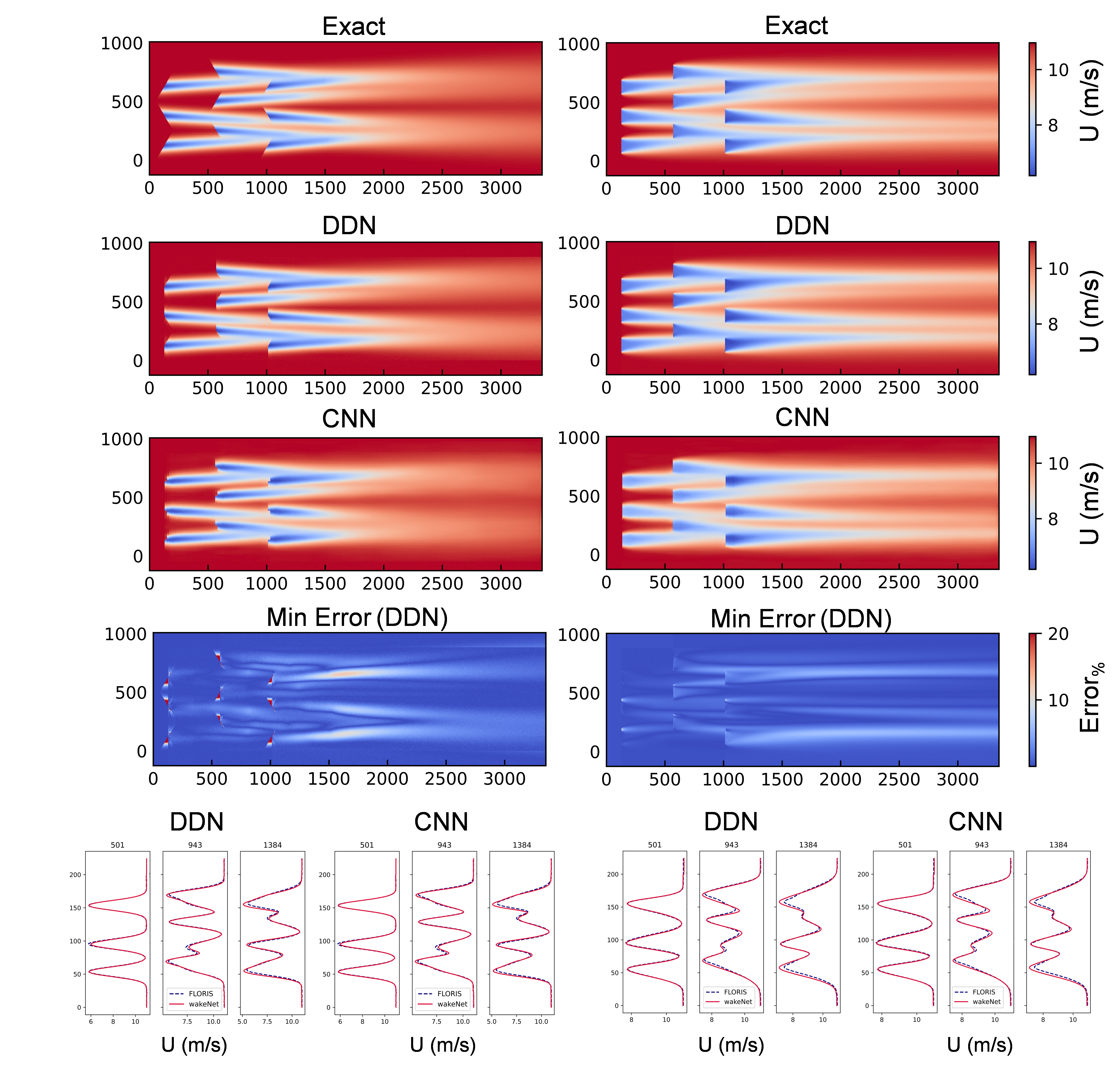}
        \caption{Example multiple wake comparison for Gaussian (left) and Curl (right) models.}
        \label{fig:Farm_wake_results}
    \end{figure}

    Figures \ref{fig:Single_wake_results} and \ref{fig:Farm_wake_results} show a single and a multiple turbine wind farm indicative cases, respectively. The left column shows the evaluations produced after training on a 2000 Gaussian wake dataset while the right column is the evaluations after training on a 2000 wake Curl dataset. The absolute relative error (\%) between the analytical and neural results is shown in Figures \ref{fig:Single_wake_results} and \ref{fig:Farm_wake_results}, and is given by:
    
    \begin{equation}
    \text{Error}_{\%} = \left | \frac{u_{\text{WakeNet}} - u_{\text{FLORIS}}}{u_{\text{stream}}} \right | \times 100.  
    \end{equation}
    
    For the single wake case (Fig. \ref{fig:Single_wake_results}), WakeNet is able to reproduce the exact solution from FLORIS for both wake models (Gaussian and Curl), with an average error of less than 2\% for the Gaussian model and less than 1\% for the Curl model, while the velocity profiles across three $y$-transects along the horizontal cross-sections of the physical domain agree with the exact transects.
    
    The multiple wake case (Fig. \ref{fig:Farm_wake_results}) selected tests the performance of WakeNet on a dense wind farm configuration with high turbine yaw settings. As expected, the absolute error increases (up to 10\%) in certain regions of the wake velocity domain (Fig. \ref{fig:Farm_wake_results}c), since for this case the superposition method also affects the results. However, the mean absolute error is less than 5\%, and the $y$-transects for both the Gaussian and Curl models in Figure \ref{fig:Farm_wake_results}d agree well with only a few exceptions, mainly in regions where multiple wakes are superimposed.
    
\subsection{TI and Power networks results}

\begin{figure}[H]
        \centering
        \subfigure[\label{fig:fcnn_power_ti_a}]{\includegraphics[height=0.32\textwidth]{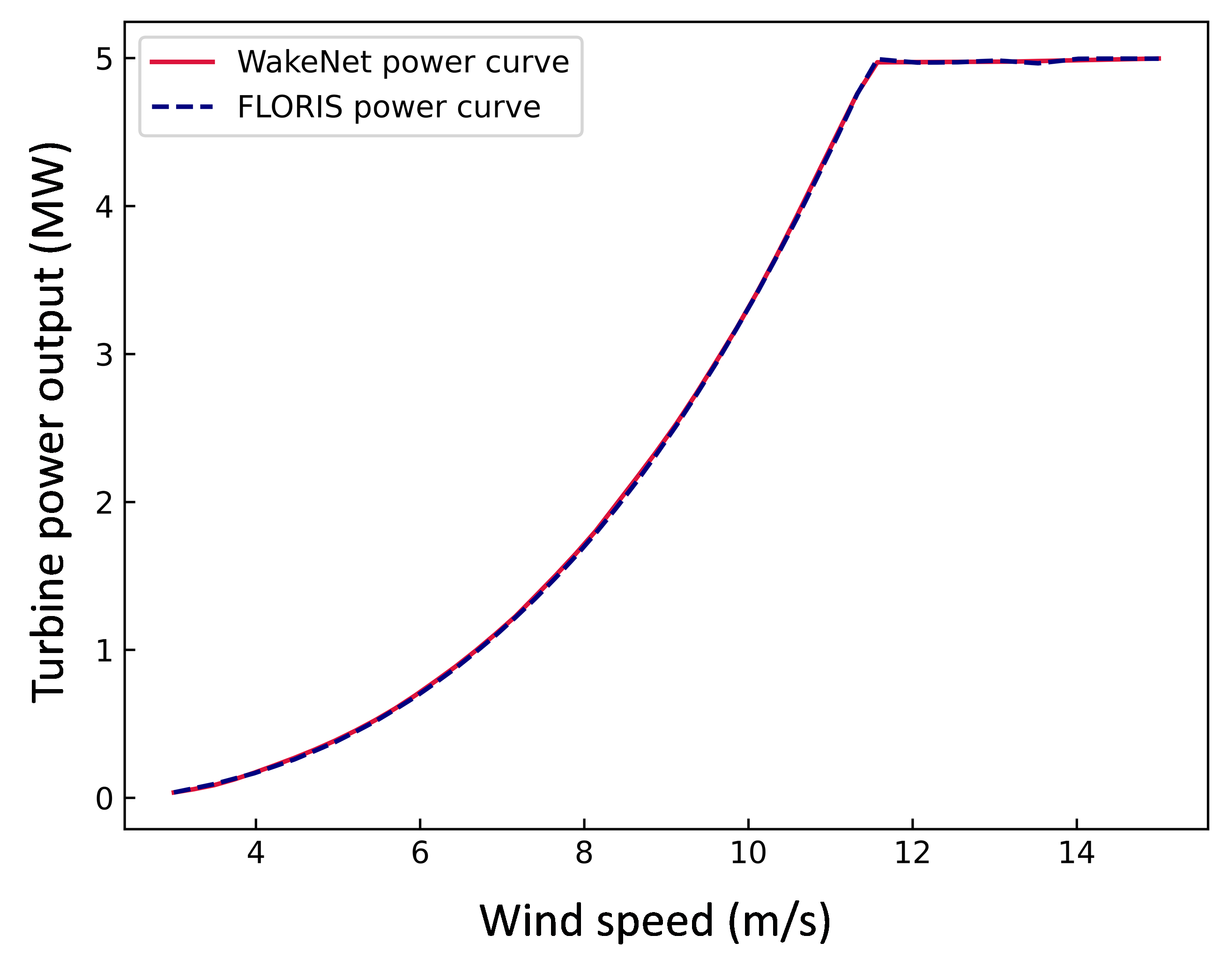}}
        \subfigure[\label{fig:fcnn_power_ti_b}]{\includegraphics[height=0.32\textwidth]{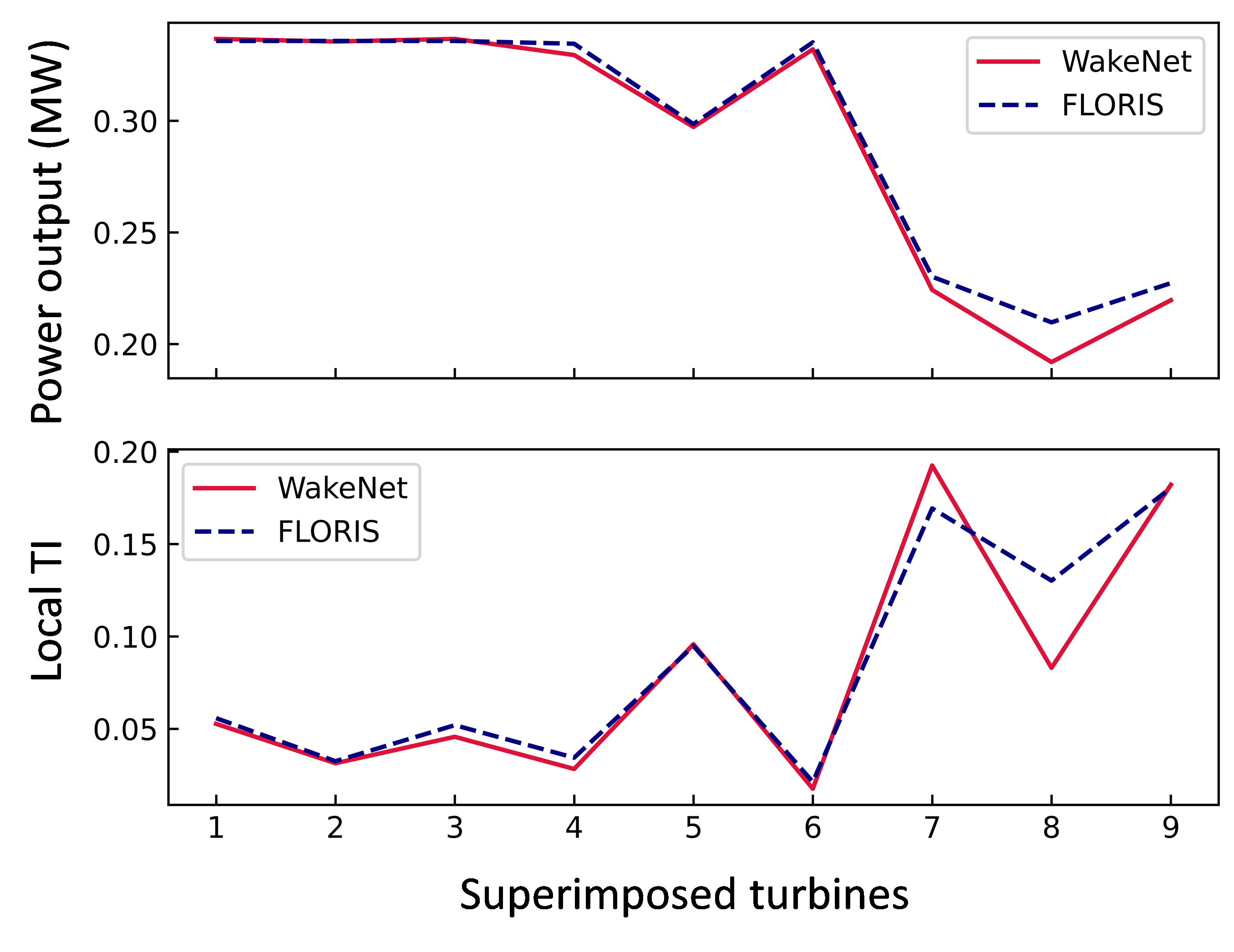}}
        \caption{WakeNet Power and TI predictions to Gaussian FLORIS comparison; (a) turbine power curve; (b) wind farm predictions against turbine number (ordered by ascending x location, if turbines have the same x location, they are ordered by ascending y location).}
        \label{fig:fcnn_power_ti}
    \end{figure}

	The power generated by every turbine and the local TI at the turbine hub is calculated using the FCNNs as described in section \ref{sec:Ti and Power networks}. Figure \ref{fig:fcnn_power_ti_a} shows the power curve for a single turbine generated by FLORIS and the predicted power by the FCNN. The mean error of the power FCNN is 1.17\% demonstrating its ability to accurately predict the power output. Figure \ref{fig:fcnn_power_ti_b} compares the power and local TI predictions of the FCNNs with FLORIS for the wind parks shown in Figure \ref{fig:Farm_wake_results}.
    The FCNN is able to accurately predict the power generated by the individual turbines, even when the turbine lies within a wake of an upstream turbine. The average power percentage error  on the test set is 2.8\%, and the local TI prediction of the FCNN has a mean TI prediction error of 7.6\% on the test data. For 200 randomly generated velocity and TI inflow conditions, the average percentage error for the power and TI prediction of the wind park shown in Figure \ref{fig:Farm_wake_results} is 3.9\% and 8.5\%, respectively.
    
\begin{figure}[H]
        \centering
        \subfigure[\label{fig:ti_enabled_wind_farm_comparison_a}]{\includegraphics[height=0.32\textwidth]{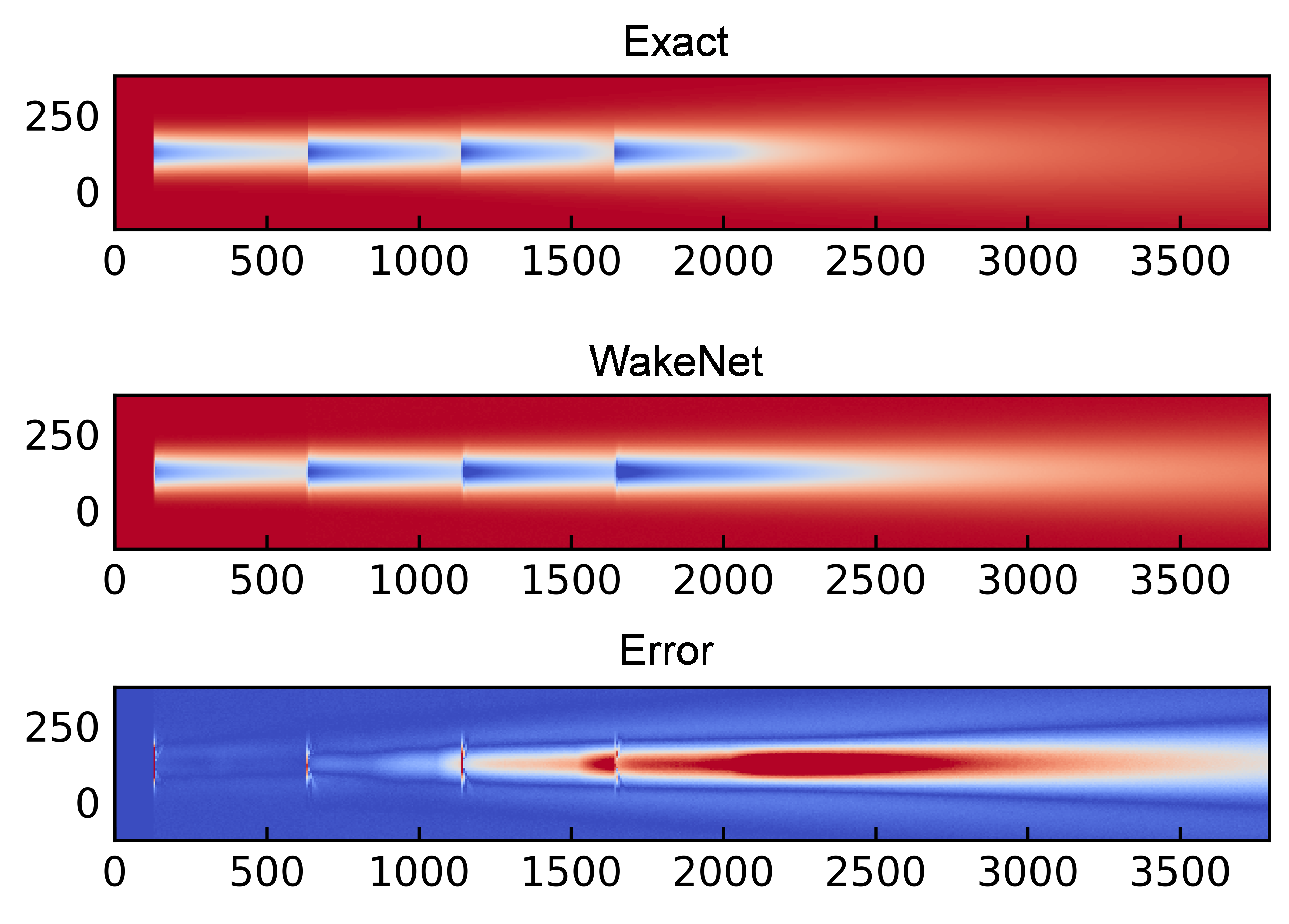}}
        \subfigure[\label{fig:ti_enabled_wind_farm_comparison_b}]{\includegraphics[height=0.32\textwidth]{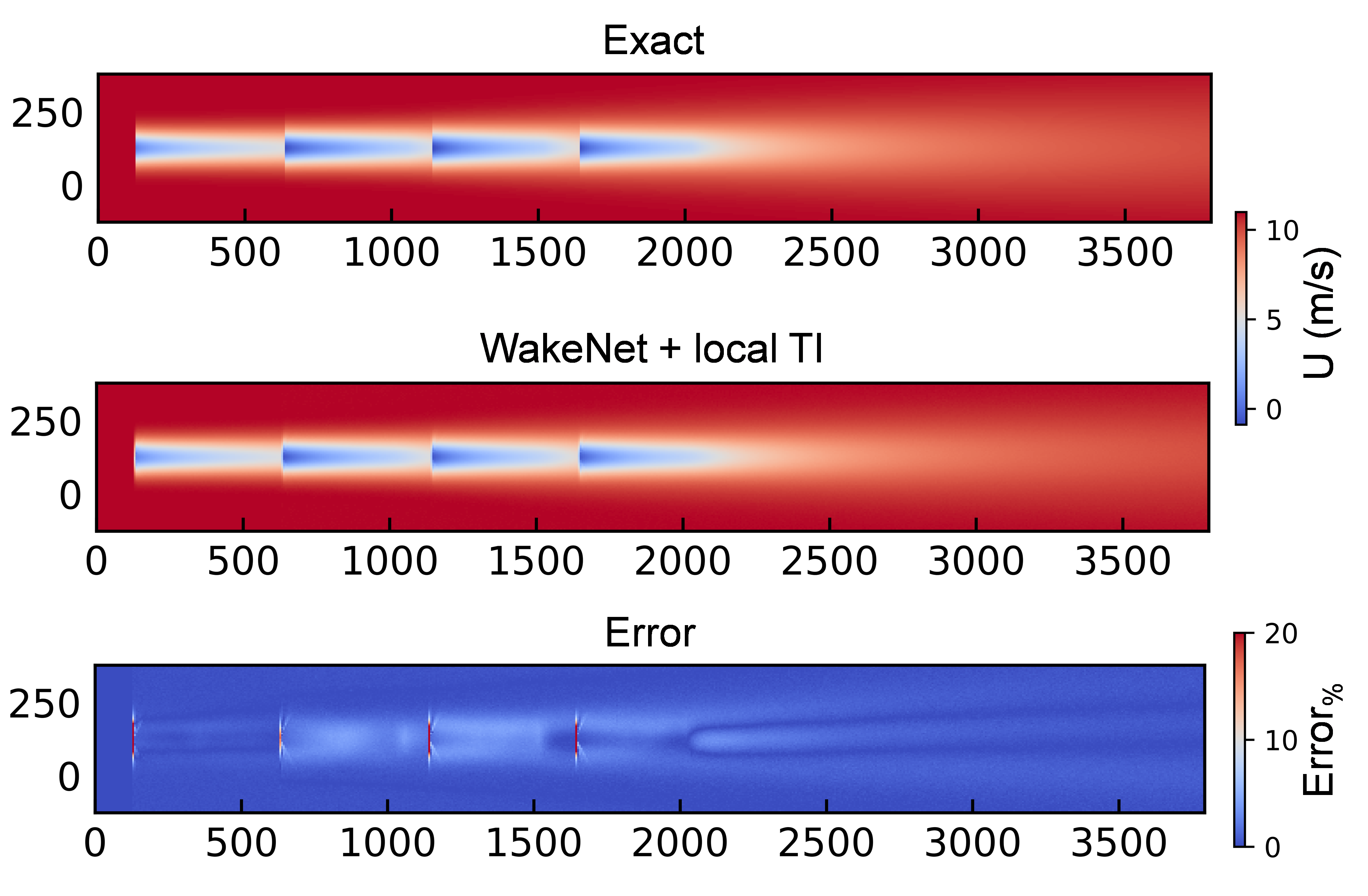}}
        \caption{Wind farm wake prediction using Gaussian DDN of four sequential turbines with no TI network (a) and with the TI network enabled (b).}
        \label{fig:ti_enabled_wind_farm_comparison}
    \end{figure}

    The concept of correcting for the TI value fluctuation downstream is illustrated in Figure \ref{fig:ti_enabled_wind_farm_comparison}, where four sequential turbines are separated by a very short distance of 2.5 wind turbine diameters. The neural network model without the TI prediction network gives a systematic error in the velocity deficit after each turbine, which leads to a 30\% error within the last wake evaluation (Fig. \ref{fig:ti_enabled_wind_farm_comparison_a}), while in the corrected case the TI network is able to maintain the error at a constant level under 3\%.
    
\begin{figure}[H]
        \centering
        \subfigure[\label{fig:power_yaw_compare_a}]{\includegraphics[width=0.4\textwidth]{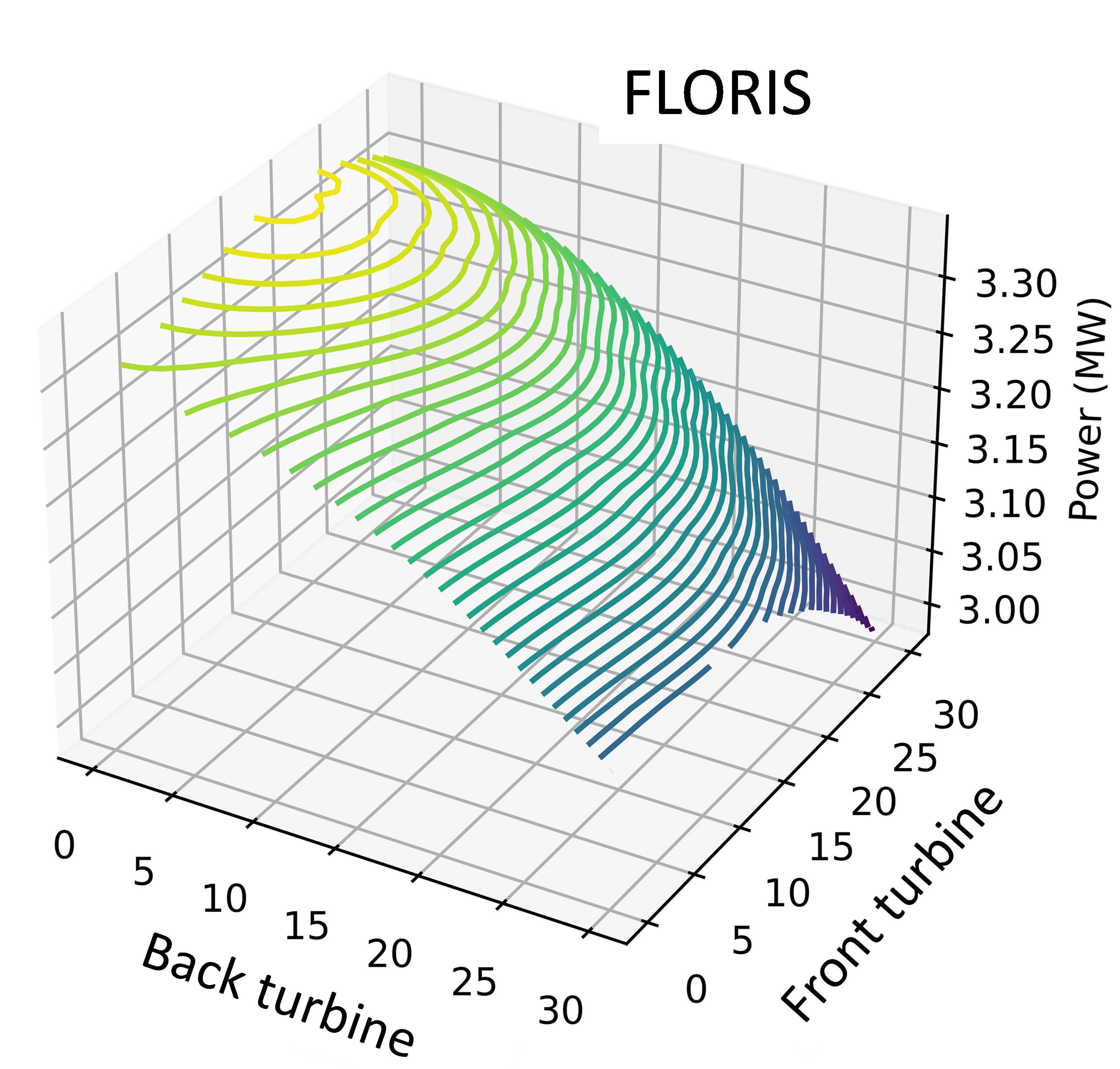}}
        \subfigure[\label{fig:power_yaw_compare_b}]{\includegraphics[width=0.4\textwidth]{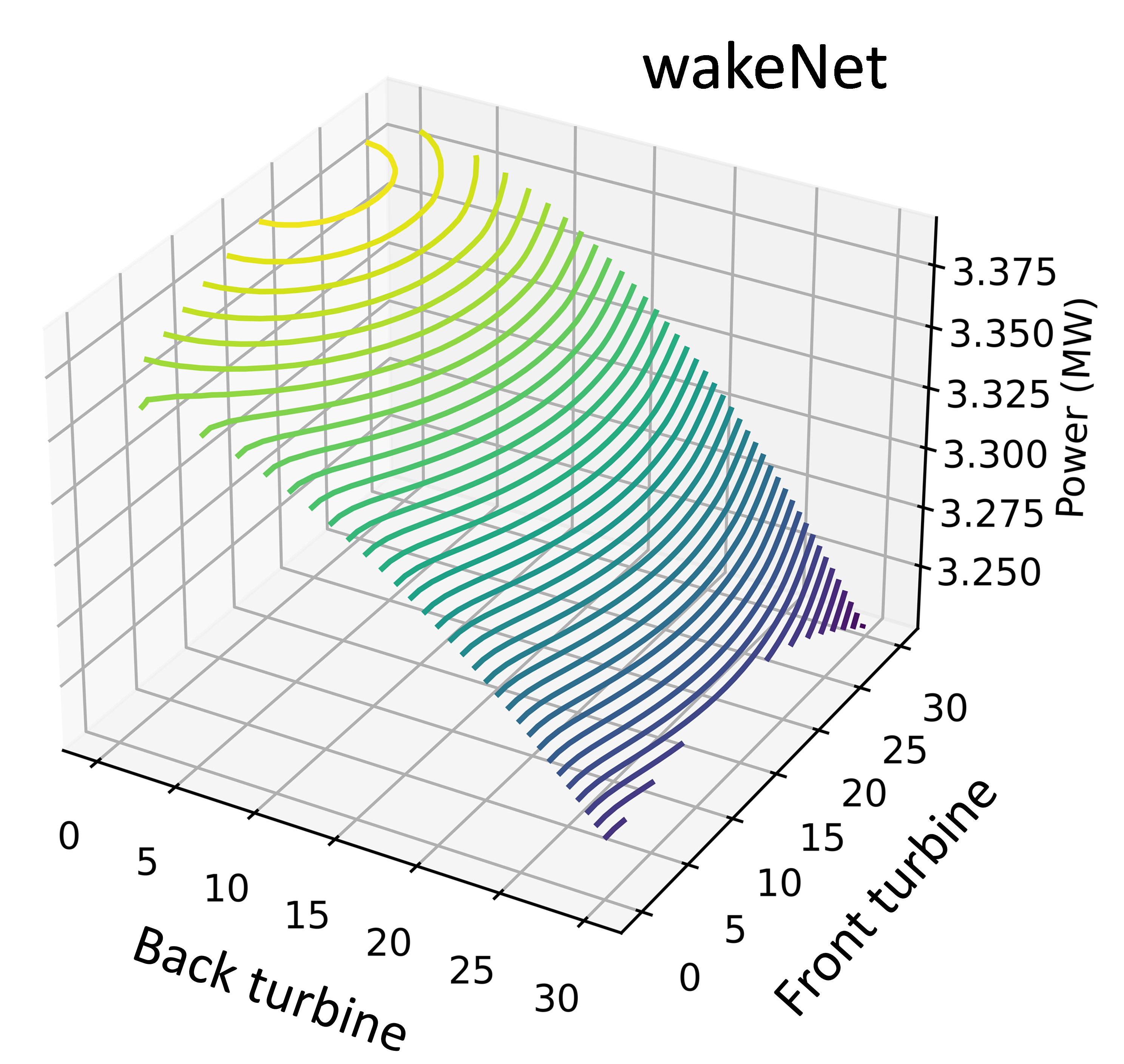}}
        \caption{Total power produced by two turbines 5D apart in the downstream direction, while their yaw varies; predictions of the neural network trained on Gaussian data (a) and FLORIS using Gaussian model (b).}
        \label{fig:power_yaw_compare}
    \end{figure}

    Finally, the correlation of the power-gain contours produced by placing two turbines 5D apart in the downstream direction and varying their yaw angles is shown in Figure \ref{fig:power_yaw_compare}. Before introducing the TI and power prediction networks, the WakeNet contour does not match with the exact solution produced by FLORIS (not shown for brevity), but with the corrections of the TI and power network predictions, both the maximum power value and angular phase space position match, namely at $\sim$3.3 MW with 15 and 0 degrees of yaw (front and back turbines, respectively).
    
\subsection{Multi-fidelity transfer learning wake results} \label{sec:mf-tl}

\begin{figure}[H]
        \centering
        \subfigure[\label{fig:DNN curl training loss/acc A}]{\includegraphics[height=0.28\textwidth]{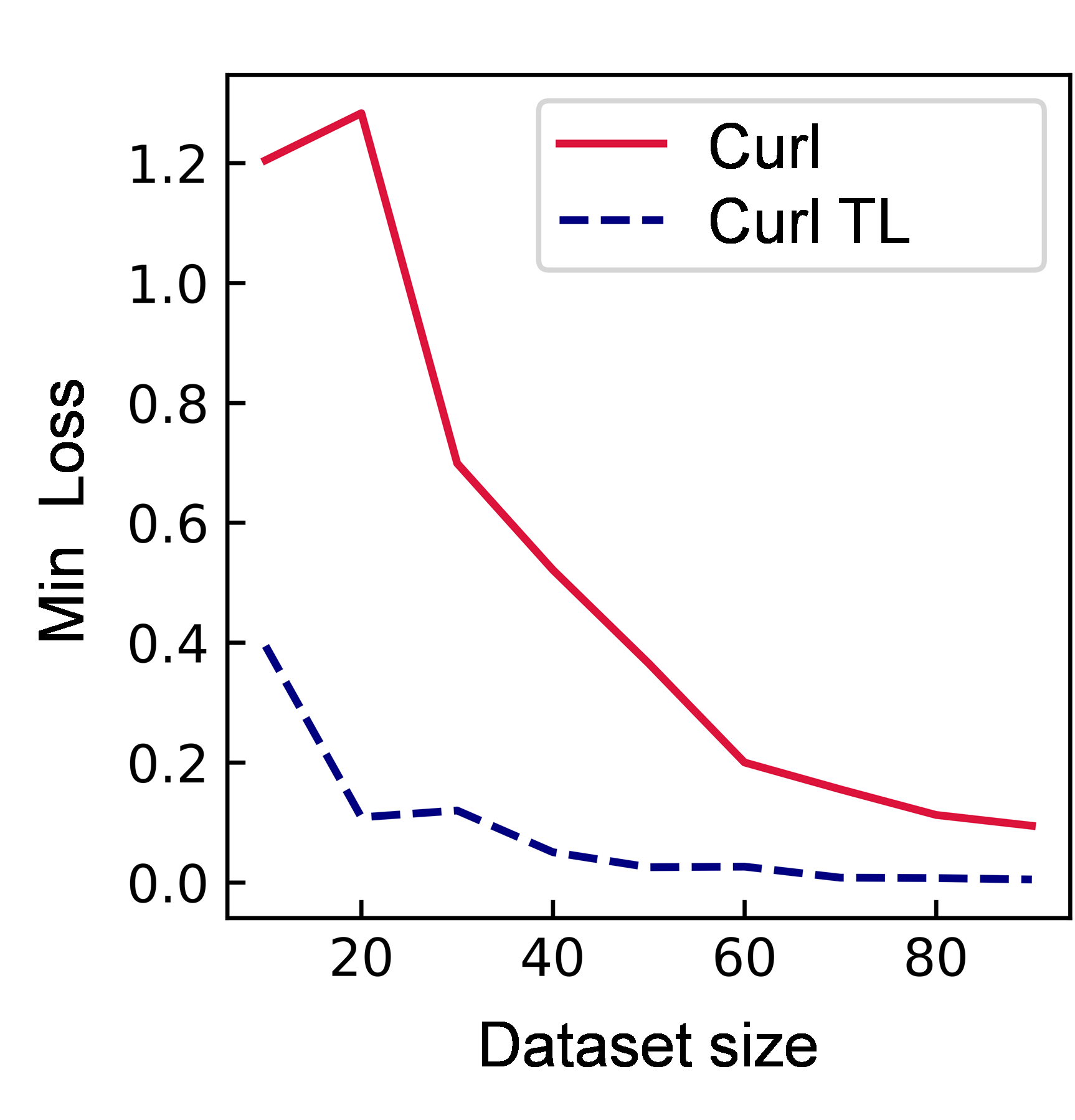}}
        \subfigure[\label{fig:DNN curl training loss/acc B}]{\includegraphics[height=0.28\textwidth]{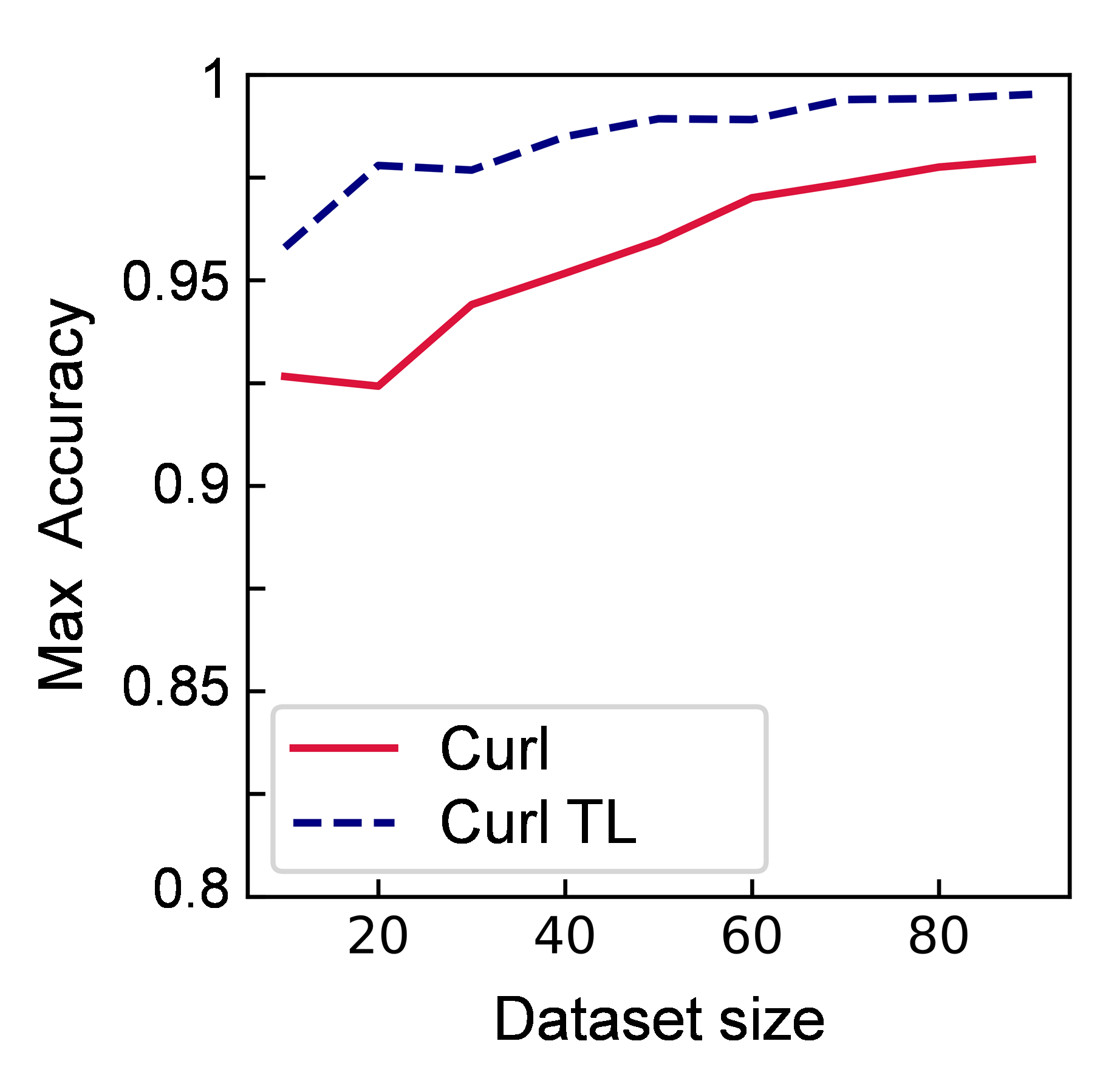}}
        \caption{Validation loss and validation accuracy plots of DDN training using the Curl model with (a) and without (b) transfer learning.}
        \label{fig:DNN curl training loss/acc}
    \end{figure}

    The advantage of using a multifidelity transfer learning approach is shown in Figure \ref{fig:DNN curl training loss/acc}, where the accuracy of different training dataset sizes on the model is compared between using only the Curl data or the transfer learning approach. In Figure \ref{fig:DNN curl training loss/acc}, each curve represents the minimum loss (Fig. \ref{fig:DNN curl training loss/acc A}) or the maximum accuracy (Fig. \ref{fig:DNN curl training loss/acc B}) achieved during training while varying the Curl dataset size from 20 to 100 wakes. The pretrained model achieves an accuracy of over 99.8\% (99.5\% for the CNN and 99.8\% for the DDN) when using only 100 wakes. An important observed characteristic of the training was that most trained models with accuracies lower than 99\% have either noisy wind fields or inaccurate free stream velocities, both of which would render the resulting wakes not suitable for participation in wind farm optimisations. For a genetic optimisation process involving the production of thousands of candidate turbine wakes, an accuracy of at least 99\% was required in order for the solutions to converge within a reasonable amount of time. If noise is present, an optimisation process might not converge or might take significantly more iterations than that of the FLORIS optimiser to obtain an optimal solution of the yaw settings or the turbine placement. A final note is that generation of a Curl dataset of 2000 wakes requires over 7 hours to compute on a Ryzen 9 5900HX processor. Since only 100 wakes are required to achieve an accuracy of over 99.8\%, using transfer learning reduces that computational cost by a factor of 20. A full CFD dataset of at least 2000 wakes would require multiple days to produce, thus an order of magnitude reduction of computational time would significantly improve the ability to produce machine learning optimisation frameworks based on even higher fidelity wake results.

\begin{figure}[H]
        \centering
        \includegraphics[width=0.80\textwidth]{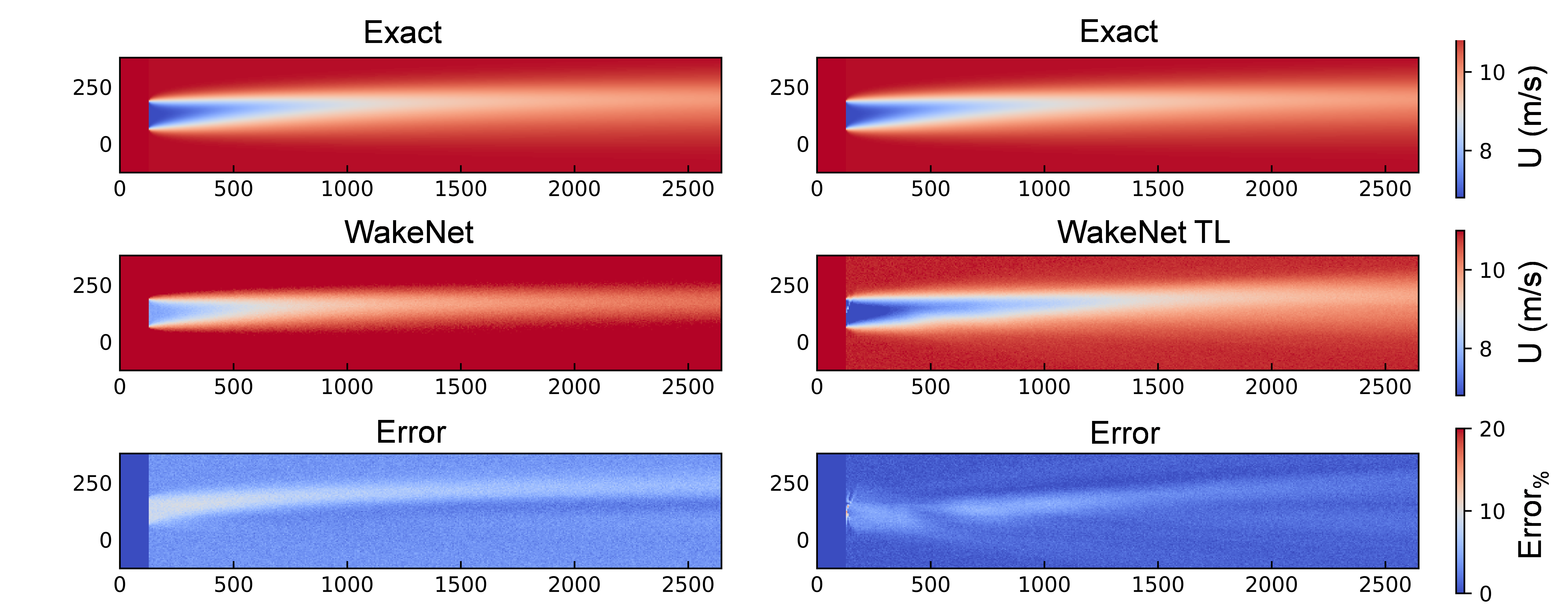}
        \caption{Indicative wake comparison using a limited Curl dataset of 100 wakes before (left) and after TL (right).}
        \label{fig:limited_curl_data_comparison}
    \end{figure}

    Figure \ref{fig:limited_curl_data_comparison} shows qualitatively, the advantage of transfer learning, (where WakeNet is trained on a reduced Curl dataset of 100 wakes) and the improvement in accuracy on a single turbine wake domain from using transfer learning is clear. The WakeNet transfer learning model is able to capture a significantly improved field, both for the velocity decrease within the turbine wake and the free stream domain. Moreover, the WakeNet model without transfer learning is not accurate enough to produce meaningful results for wind farm total power evaluation, meaning it is inappropriate for optimisation scenarios.

\subsection{Computational time scaling}\label{sec:computational_time_scaling}
\begin{figure}[H]
    \centering
    \subfigure[\label{fig:time_scaling_a}]{\includegraphics[height=0.32\textwidth]{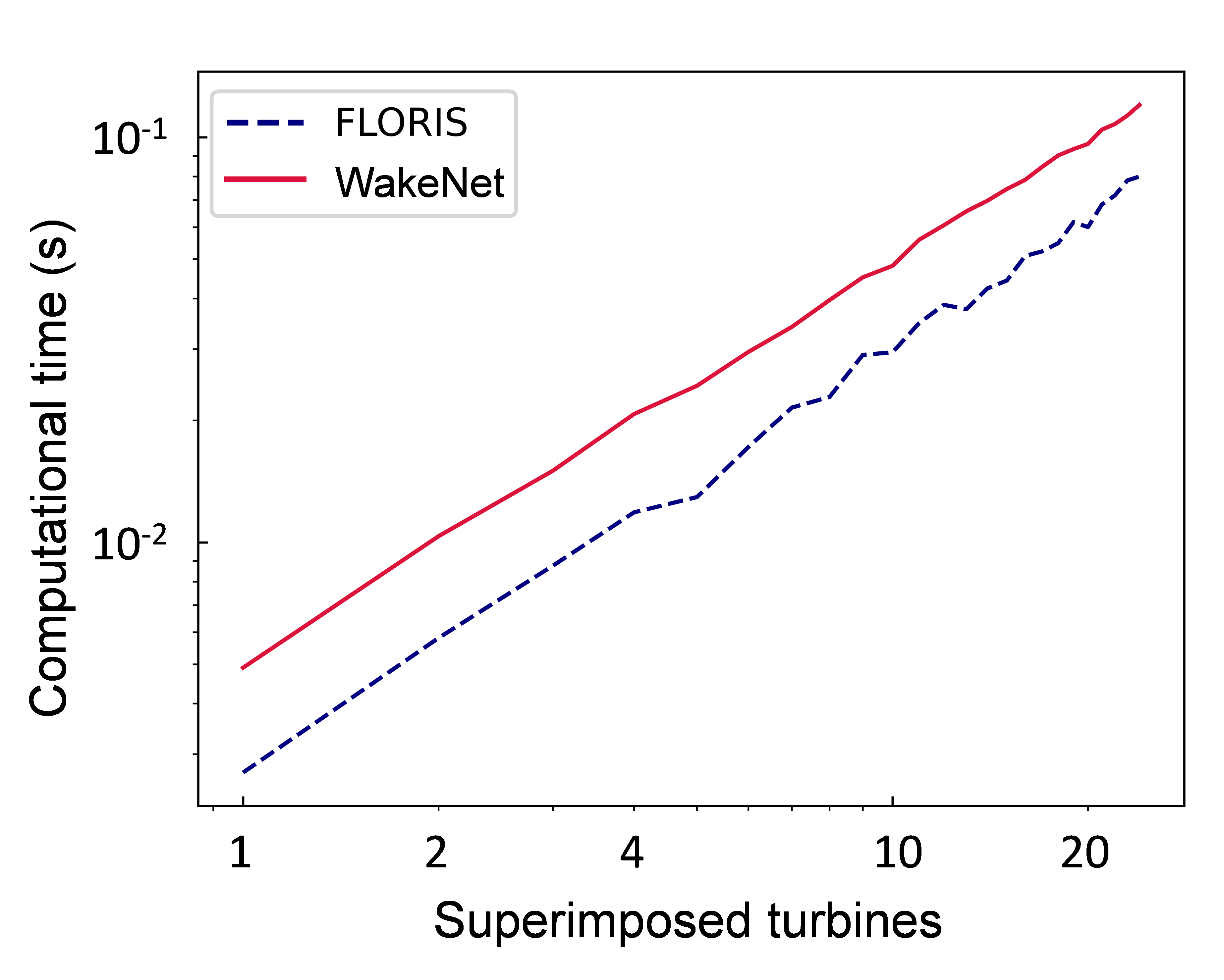}}
    \subfigure[\label{fig:time_scaling_b}]{\includegraphics[height=0.32\textwidth]{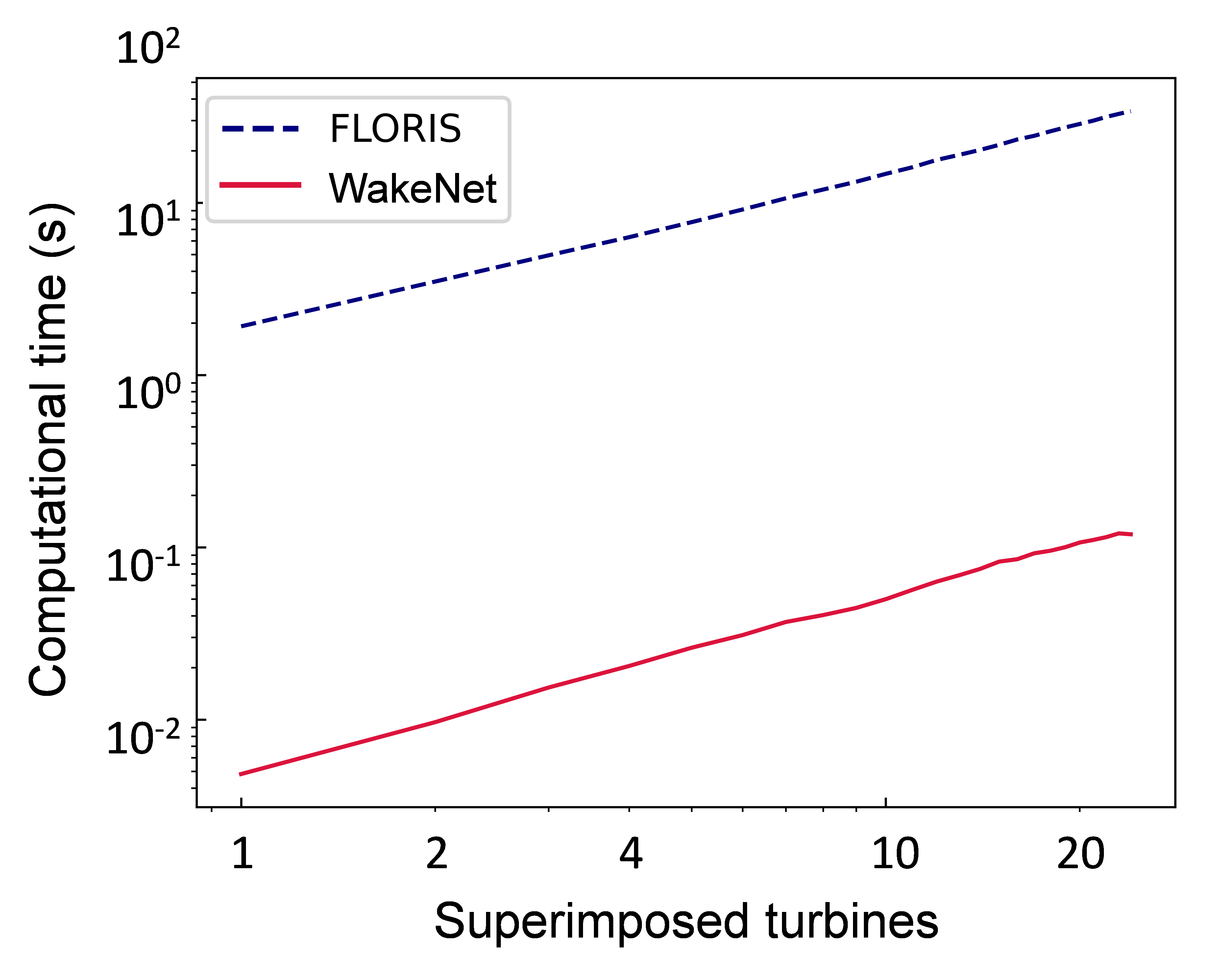}}
    \caption{Logarithmic computational time scaling for the simulation of up to 24 turbines using Floris and WakeNet for the Gaussian (a) and Curl (b) models.}
    \label{fig:time_scaling}
    \end{figure}

    The performance of WakeNet is further tested in terms of the computational time scaling for the superposition of up to 24 turbines. Fig. \ref{fig:time_scaling_a} shows the log-log time comparisons between FLORIS and WakeNet for the Gaussian model and Fig. \ref{fig:time_scaling_b} shows the same for the Curl models. As expected, for the Gaussian model, both WakeNet and FLORIS scale similarly, and WakeNet offers no computational time gains as both FLORIS and WakeNet can compute/evaluate a 24-turbine wind farm in under one second. However, for the Curl model, WakeNet is able to evaluate the wake field more than two orders of magnitude faster (0.5 s vs 40 s for the 24-turbine case). As before, the rate at which the computational time increases per superimposed turbine is the same for both WakeNet and FLORIS. We note that the computational times include the time take taken to calculate the total wind farm power. However, to do this, WakeNet is also producing the 2D velocity field array that can easily be visualised with no significant time cost. FLORIS does not do this, and if the FLORIS computational time included the process to generate this 2D velocity field array in FLORIS, the computational time gains would be significantly higher, even for the Gaussian wake case.

    A similar performance is achieved by the CNN module and therefore, for simplicity reasons, the DNN module is selected for the remaining results presented in this study.
    
\subsection{Optimisation} 
\begin{figure}[H]
        \centering
        \includegraphics[width=0.9\textwidth]{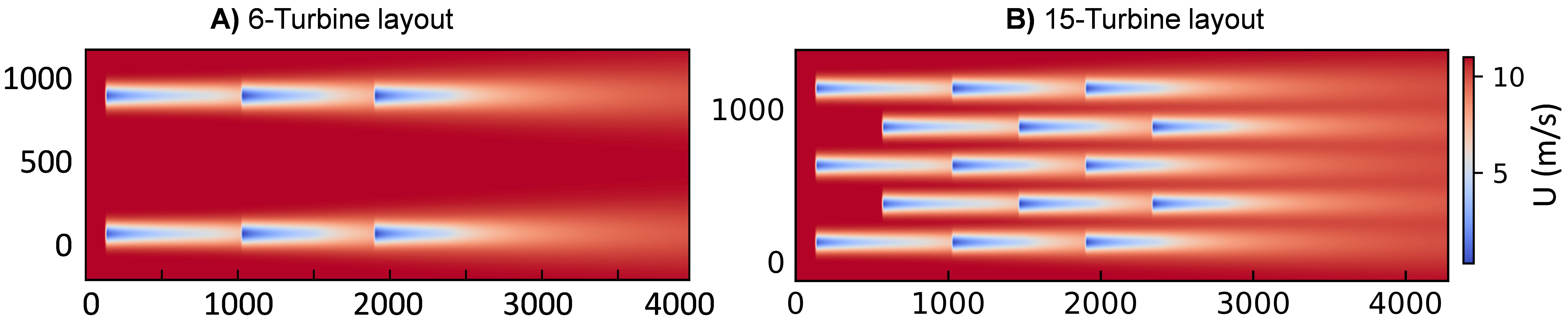}
        \caption{Layouts used for each optimisation case.}
        \label{fig:farm_layouts_optimisation}
    \end{figure}

    The yaw angle and layout optimisation scenarios considered in this study are performed on two different wind farm layouts, one 6-turbine layout (case A) and one dense 15-turbine layout (case B), as shown in Figure \ref{fig:farm_layouts_optimisation}. This work uses SciPy’s SLSQP optimiser \cite{2020SciPy-NMeth} to optimise the total power generated by the wind farm. The exact power output produced for the initial configurations, the optimal yaw settings and the optimal turbine positions in the domain are always calculated by FLORIS so that the performance between WakeNet and FLORIS can be compared.
    
\subsubsection{Gaussian-based optimisation}
    
\begin{figure}[H]
        \centering
        \includegraphics[width=0.85\textwidth]{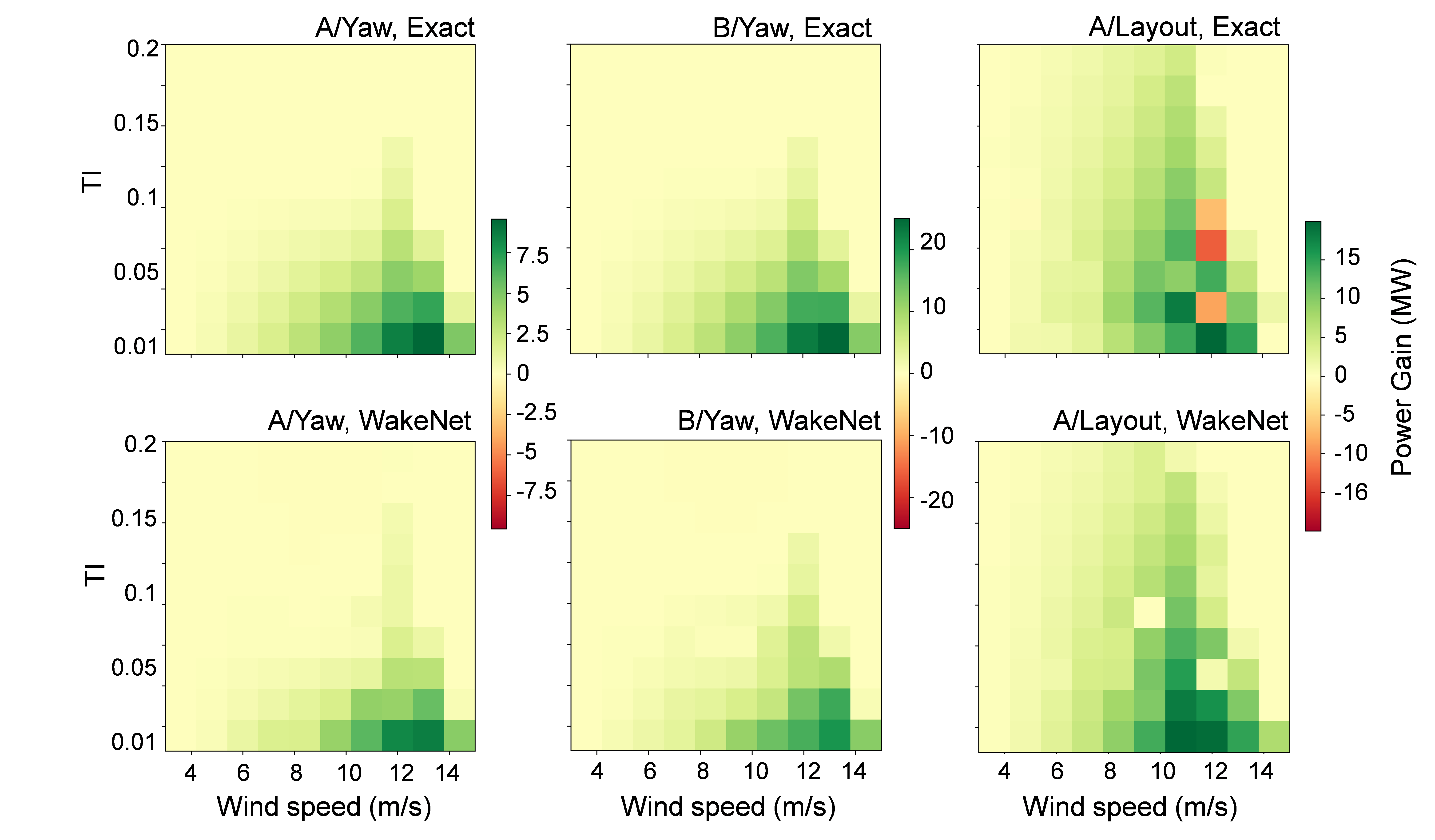}
        \caption{Yaw and layout optimisation power gain heatmap comparisons using WakeNet trained on a dataset of 2000 Gaussian wakes.}
        \label{fig:gaussian_heatmap}
    \end{figure}

    In order to assess the capability of the neural network to obtain the optimal power gain, Figure \ref{fig:gaussian_heatmap} shows the corresponding total farm power heatmaps for the Gaussian-based optimisation. The heatmaps can be used to compare FLORIS optimised results (Exact) with those from WakeNet after both have been used to perform yaw optimisation for both farms A and B, as well as layout optimisation on farm A, under a range of TI and wind speed values. As evident from these heatmaps, the region with the highest potential power gain is around the bottom right corner, which is defined by low TI and high inlet speeds. 
    
    For the two yaw cases, WakeNet optimisation finds optimal yaw settings which yield at least 90\% of the power gain as the FLORIS optimiser. For the layout optimisation case the overall performance of WakeNet is similar to FLORIS, with some noise visible in the power gain region produced by both optimisers. Furthermore, the average computational time cost for each optimisation heatmap is of the same order of magnitude for both FLORIS and WakeNet optimisers (Table \ref{tbl:gaussian_optimisation_table}), as expected based on the time scaling of Figure \ref{fig:time_scaling}.

    \begin{table}[H]
    \caption{Gaussian wake model average optimisation timings.}
    \centering
    \begin{tabular}{|c|c|c|c|} \hline
    \backslashbox[3cm]{Model}{Case/Type}
    & A/Yaw, Av. time (s) & B/Yaw, Av. time (s) & A/Layout, Av. time (s) \\\hline
    FLORIS & 8 & 31.5 & 7.3 \\\hline
    WakeNet & 5.2 & 39.5 & 5 \\\hline
    \end{tabular}
    \label{tbl:gaussian_optimisation_table}
    \end{table}

\subsubsection{Curl-based optimisation}

\begin{figure}[H]
        \centering
        \includegraphics[width=0.6\textwidth]{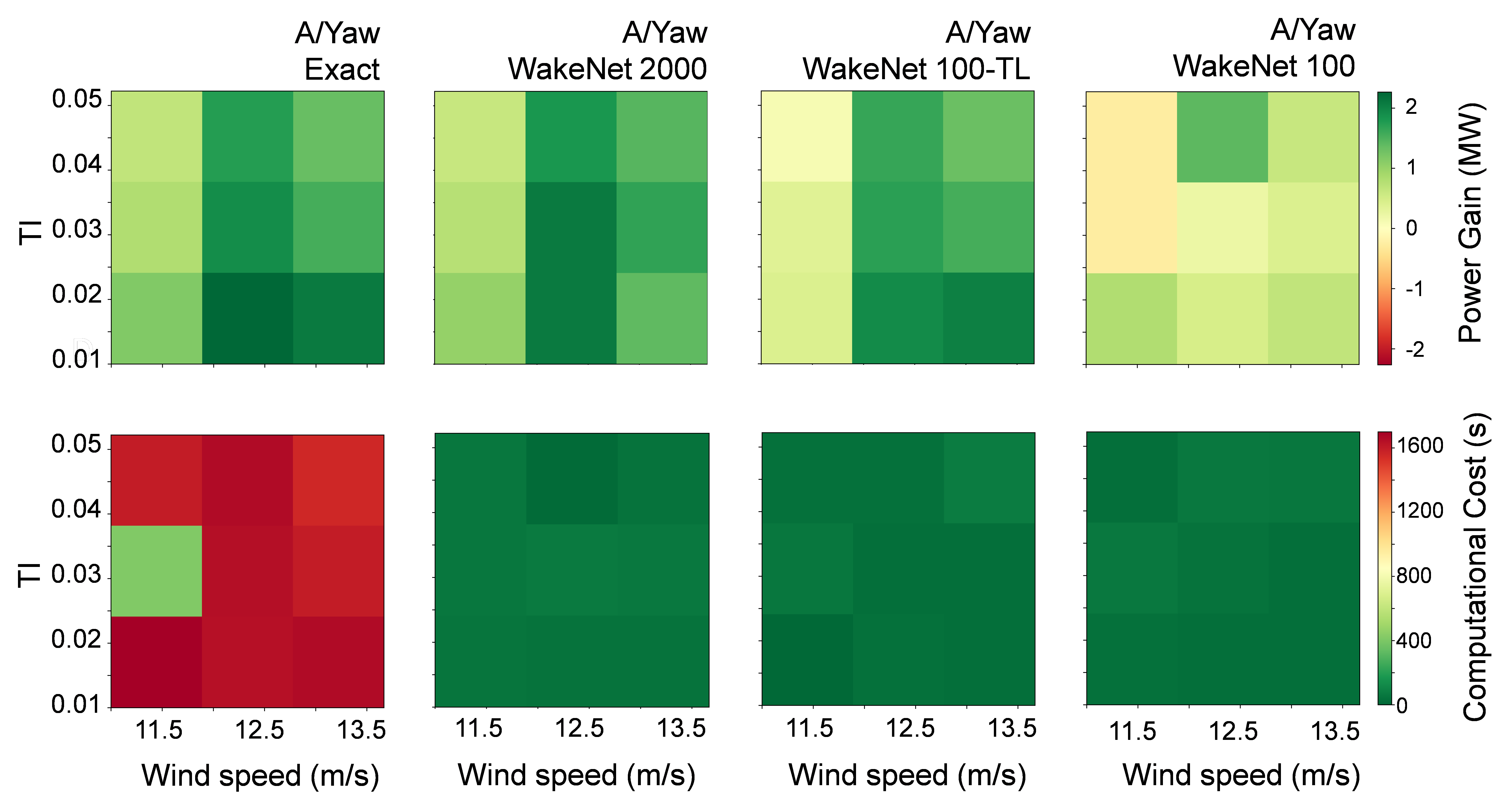}
        \caption{Yaw optimisation heatmaps. A comparison between the power gain and the computational cost between the exact solution and WakeNet with Curl datasets of 2000 wakes, 100 wakes with TL and 100  wakes without TL.}
         \label{fig:ddn_3.2_1}
    \end{figure}

    The Curl-based WakeNet is used for producing similar optimisation heatmaps for layout A. Due to the high computational demands of the Curl model, the Curl-based optimisation was performed within the range of high potential power gain regions that were previously identified by the Gaussian-based optimisation heatmaps. Figure \ref{fig:ddn_3.2_1} shows that the optimisation produced by FLORIS exact solutions has a very high correlation with both WakeNet (trained on 2000 Curl wakes) and WakeNet TL (trained on 100 Curl wakes), where both neural network models provide optimised yaw angles which yield more that 90\% of the potential power of the exact solution. The WakeNet model trained on 100 wakes without TL, however, produces less than 30\% of that power gain, which is expected based on the lower training accuracy and the qualitative results of section \ref{sec:mf-tl}.
    
    Nonetheless, the significance of these results lies in the corresponding computational time heatmaps and their time averaged values in Table \ref{tbl:curl_average_optimisation_timigs}. While FLORIS requires 25 minutes on average for each optimisation (pixel) of the presented heatmaps, WakeNet requires 35 seconds on average.

    \begin{table}[H]
    \centering
    \caption{Curl average yaw optimisation timings.}
    \begin{tabular}{|c|c|} \hline
    \backslashbox[3cm]{Model}{Case/Type}
    & A/Av. time (s) \\\hline
    FLORIS & 1500 \\\hline
    WakeNet 100 & 38 \\\hline
    WakeNet 100-TL & 32 \\\hline
    WakeNet & 41 \\\hline
    \end{tabular}
    \label{tbl:curl_average_optimisation_timigs}
    \end{table}

\begin{figure}[H]
        \centering
        \subfigure[\label{fig:optimised_curl_layouta}]{\includegraphics[width=0.9\textwidth]{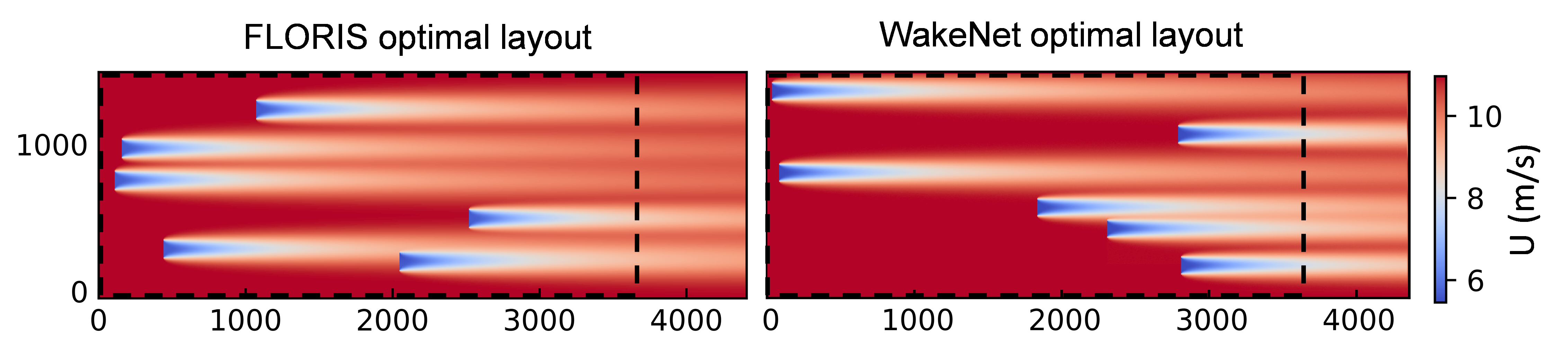}}
        \subfigure[\label{fig:optimised_curl_layoutb}]{\includegraphics[width=0.85\textwidth]{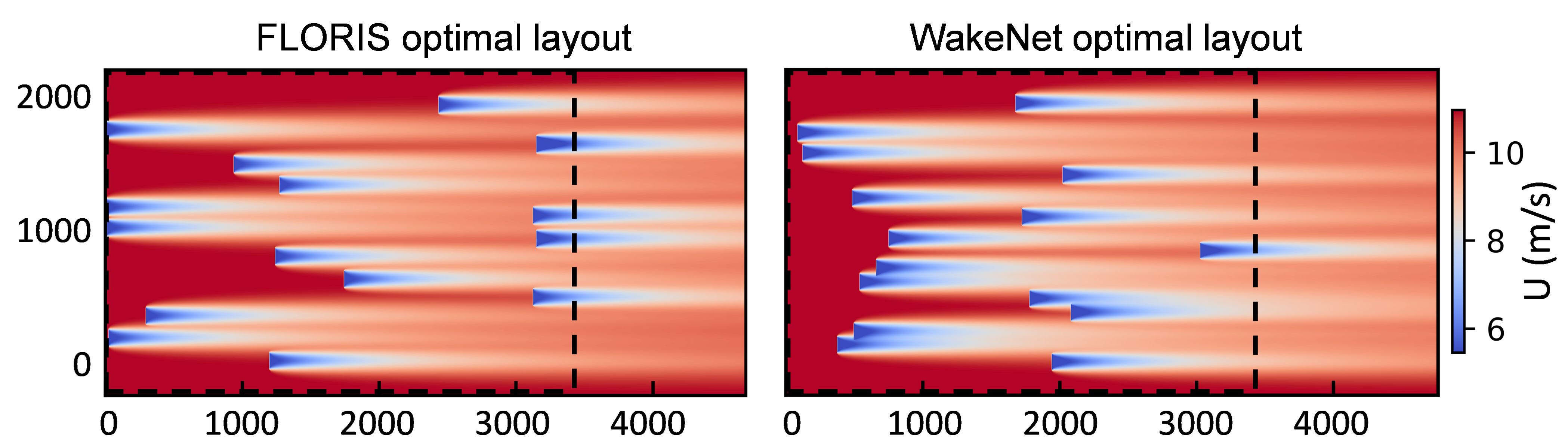}}
        \caption{Curl-based optimised layout produced by FLORIS and WakeNet optimisers for 6-turbine (a) and 15-turbine wind farms (b). The dashed line indicates the spatial constraint.}
        \label{fig:layout_opts}
    \end{figure}

    Moreover, two indicative layout optimisations are performed for both layouts A and B of Figure \ref{fig:farm_layouts_optimisation}, with a fixed wind speed of 11 m/s and ti 0.05 (Table \ref{tbl:curl_information_table}). The optimal layouts of FLORIS and WakeNet optimisers are shown in Figure \ref{fig:layout_opts}. Even though the optimal solution is not unique (hence the different converged layouts between the two optimisers), the adopted strategy is similar: the turbines are spread in the downstream and vertical directions to minimise blockage losses. Based on the optimisation results of Table \ref{tbl:curl_information_table}, the WakeNet optimiser captures 3.5\% out of 4.4\% and 60.83\% out of 61.28\% of the potential power gain percentage obtained by FLORIS. To provide these power gains for layout A \ref{fig:optimised_curl_layouta}, WakeNet requires less than a minute compared to the two hours required by FLORIS. For layout B, the time required by FLORIS optimizer is 36 hours compared to 13 minutes required by WakeNet optimiser. Hence the total computational time gains are higher than two orders of magnitude, which is consistent with the computational gains in the time-scaling study of section \ref{sec:computational_time_scaling}. Note that the layout optimisation time grows exponentially as the number of turbines increases due to the increasing combinations of wake interactions.
    
    In summary, we believe that the computational time gains presented for the Curl-based optimisation scenarios constitute an important stepping stone towards real-time yaw optimisation and more complex layout optimisation, e.g. optimisation under uncertainty, using high-fidelity wake models.

    \begin{table}[H]
    \centering
    \caption{Curl layout optimiser information.}
    \begin{tabular}{|c|c|c|c|c|} \hline
    \backslashbox[3cm]{Info}{Case/Model}
    & A/FLORIS & A/WakeNet & B/FLORIS & B/WakeNet\\\hline
    {Initial Power (MW)} & {4.06} & {4.06} & {42.2} & {42.2} \\\hline
    {Power Gain (\%)} & {47.1} & {46.35} & {61.28} & {60.83} \\\hline
    {Comp. Cost (min)} & {121} & {0.5} & {2160} & {13} \\\hline
    \end{tabular}
    \label{tbl:curl_information_table}
    \end{table}
    
\section{Conclusions}

    The novel ML framework WakeNet, can reproduce generalised 2D turbine wake fields at hub-height over a wide range of yaw settings, wind speeds and turbulence intensities, with a mean accuracy of 99.8\% compared to the downstream velocity domain produced by either the Gaussian or Curl models of FLORIS. Two FCCNs are deployed to approximate the complex 3D quantities of local TI and generated power from the 2D flow predictions of WakeNet. These additional networks are capable of predicting the local TI and power values with a mean accuracy of 98\%. All of the ML modules making up WakeNet are validated through multiple superposition examples.
    
    The computational cost scaling of WakeNet as the number of superimposed turbines increases is of the same order of magnitude as FLORIS with the Gaussian wake model. However, when WakeNet is trained on the more sophisticated Curl wake model, the computational time gains become two orders of magnitude higher that FLORIS with the Curl wake model. The wake evaluation time of WakeNet when trained on ever higher-fidelity CFD dataset is expected to be similar, thus further increasing the computational time gains. This will be the topic of future research. Furthermore, the robustness and overall performance of WakeNet on various yaw and layout optimisation scenarios across a range of wind speeds and TIs has been validated through power-gain heatmaps. When trained on the Gaussian model, WakeNet is able to reproduce similar optimal configurations (obtaining at least 90\% of power gained by FLORIS optimisation), at a similar order of computational cost, as expected. However, when considering the Curl model, WakeNet provides similar power gains at least two orders of magnitude faster than FLORIS on average, which is also indicated by the computational-cost scaling results.

    Finally, multi-fidelity transfer learning has been deployed for fine-tuning the network weights that have been pre-trained on the low-fidelity wake model (Gaussian dataset of 2000 wakes) to obtain accurate wake results for a high-fidelity wake model using a limited dataset (Curl dataset of 100 wakes). The trained transfer learning network has been deployed for indicative yaw and layout optimisation, where it finds the optimal configurations two orders of magnitude faster than FLORIS on average, accelerating optimisations that took 1--2 hours down to a few seconds.

    These promising results show that generalised wake modelling with machine learning tools can be accurate enough to contribute towards active yaw and complex layout optimisation applications. Furthermore, the neural network models can be trained on high-fidelity wake models to produce more realistic optimised configurations at a fraction of the computational time, rendering real-time applications in active yaw optimisation possible. Multi-fidelity transfer learning techniques can be useful in producing similar flow and optimisation results using limited high-fidelity datasets, when the solution times would not allow for the creation of a large dataset (e.g. 2000 wakes produced by LES solvers is a computationally extremely expensive task). The proposed methodology could enable  maximising wind farm power gains at a minimal installation/operation cost. Planned future implementations include the addition of extra network inputs predictors such as  veer and wind direction; the introduction of higher-fidelity TL steps using CFD datasets to further improve realistic wake approximations; testing of our framework on real wind-farm scenarios; parallel computing of forward evaluations for faster optimisation.

\typeout{}
\bibliographystyle{unsrt}
\bibliography{references}

\end{document}